\documentclass[fleqn,10pt]{wlscirep}
\usepackage[utf8]{inputenc}
\usepackage[T1]{fontenc}
\usepackage{booktabs}
\usepackage{array}
\usepackage{float}

\usepackage{graphicx}
\usepackage{caption}
\usepackage{subcaption}
\usepackage{afterpage}

\usepackage{color}
\usepackage{multirow}
\usepackage{xcolor}
\usepackage{tcolorbox}
\usepackage{xspace}
\newtcolorbox{graycomment}{
  colback=gray!10,
  colframe=gray!50,
  leftrule=3pt,
  rightrule=0pt,
  toprule=0pt,
  bottomrule=0pt,
  arc=0pt,
  outer arc=0pt,
  left=5pt,
  right=5pt,
  top=3pt,
  bottom=3pt
}
\newcommand{\mydataset}{\texttt{AgoraSpeech}\xspace}
\newcommand{\myitem}[1]{%\vspace{0.25\baselineskip}
\noindent\textbf{#1}\xspace}%\vspace*{0.04in}
\title{AgoraSpeech: A multi-annotated comprehensive dataset of political discourse through the lens of humans and AI} 

\author[1,*]{Pavlos Sermpezis}
\author[1]{Stelios Karamanidis}
\author[1]{Eva Paraschou}
\author[1]{Ilias Dimitriadis}
\author[1]{Sofia Yfantidou}
\author[1]{Filitsa-Ioanna Kouskouveli}
\author[2]{Thanasis Troboukis}
\author[2]{Kelly Kiki}
\author[3]{Antonis Galanopoulos}
\author[1]{Athena Vakali}
\affil[1]{Data \& Web Science Lab, School of Informatics, Aristotle University of Thessaloniki, Thessaloniki, 54124, Greece}
\affil[2]{incubator for Media Education and Development (iMEdD), Athens, 10562, Greece}
\affil[3]{School of Political Sciences, Aristotle University of Thessaloniki, 54124, Greece}
\affil[*]{corresponding author%(s)
: Pavlos Sermpezis (sermpezis@csd.auth.gr)%, \red{XXXX} (\red{xxx@xxx})
}
\begin{abstract}
Political discourse datasets are important for gaining political insights, analyzing communication strategies or social science phenomena. Although numerous political discourse corpora exist, comprehensive, high-quality, annotated datasets are scarce. This is largely due to the substantial manual effort, multidisciplinarity, and expertise required for the nuanced annotation of rhetorical strategies and ideological contexts. In this paper, we present \mydataset, a meticulously curated, high-quality dataset of 171 political speeches from six parties during the Greek national elections in 2023. The dataset includes annotations (per paragraph) for six natural language processing (NLP) tasks: text classification, topic identification, sentiment analysis, named entity recognition, polarization and populism detection. A two-step annotation was employed, starting with ChatGPT-generated annotations and followed by exhaustive human-in-the-loop validation. The dataset was initially used in a case study to provide insights during the pre-election period. However, it has general applicability by serving as a rich source of information for political and social scientists, journalists, or data scientists, while it can be used for benchmarking and fine-tuning NLP and large language models (LLMs).

\end{abstract}

\begin{document}

\flushbottom
\maketitle
%  Click the title above to edit the author information and abstract
\thispagestyle{empty}

\section*{Background \& Summary}
Analyzing political discourse is essential for uncovering rhetorical strategies, ideological nuances, and communication patterns, now increasingly facilitated by the wealth of online content and Artificial Intelligence (AI) tools. Fully leveraging these advanced techniques requires robust data for training and methods for assessing their accuracy and reliability. Developing structured and annotated benchmarking datasets could revolutionize political science, by enabling deeper insights, offering new standards for assessing efficiency of AI techniques, and revealing new opportunities or even challenges that need to be addressed by future research.

\subsection*{Political discourse datasets: limitations, challenges, and motivation} 
In the context of political science, the availability of high-quality and meticulously labeled benchmark datasets holds immense value for advancing research in political discourse analysis. While such datasets exist for various purposes such as metaphor analysis~\cite{ahrens2018using,charteris2009metaphor}, party affiliation classification~\cite{yu2008classifying}, topic identification~\cite{dilai2020automatic}, and populism detection~\cite{populismus2020} they typically suffer from limitations in the annotation quality and completeness, scale and generality~\cite{card2015media,wachsmuth2017computational}. 
For example, few existing political speech corpora offer detailed, fine-grained annotations of rhetorical strategies~\cite{ilsp2020}, emotional appeals, ideological positioning~\cite{11356/1864}, or persuasive techniques~\cite{Lehmann:2024}. Moreover, many available datasets are either too small to allow for meaningful statistical analysis or suffer from inconsistencies in transcription, annotation, or quality control\cite{di2024sampled, chen2022election2020}. These limitations hinders the reproducibility and scalability of research based on such datasets, and the depth of analysis that can be performed on the discourse.

Political speeches are a rich source of data for analyzing discourse, rhetorical strategies, ideological stances, and communication patterns across the political spectrum. Despite the abundance of political discourse available, there is a critical need for high-quality, annotated datasets that offer deep insights into the structure and substance of these speeches. 
The absence of high-quality annotations can be attributed to the substantial manual effort required for annotating large corpora of texts\cite{li2023coannotating}. Furthermore, the domain-specific expertise demanded for accurate labeling, typically possessed by journalists or political scientists, often hinders the publication of such datasets. The meticulousness required in the labeling methodology is paramount, considering the potential societal impacts, as underscored by regulatory frameworks such as the AI Act's guidelines on high-risk applications for the administration of justice and democratic processes~\cite{madiega2021artificial}.

$\rightarrow$ \textit{In light of these gaps, the primary motivation for this paper is to present a comprehensive, high-quality annotated dataset of political discourse, designed for a wide range of applications, from topic classification to populism detection.}

\subsection*{LLMs as data annotators: effectiveness, opportunities and the need for benchmarking}
Given the need for annotated datasets, why not leverage AI technologies to automate the annotation of existing corpora of public discourse?

In recent years, the landscape of Natural Language Processing (NLP), transformed by the emergence of Large Language Models (LLMs) like ChatGPT, has expanded well beyond its conventional research applications~\cite{brown2020language}. LLMs can help augment human intelligence, from laymen crafting witty social media posts or eloquent cover letters to businesses streamlining customer interactions and developing engaging marketing content. 
Several studies have explored LLMs’ potential to serve as an annotator for textual data, offering insights into different NLP tasks~\cite{gilardi2023chatgpt,zhu2023can}. For example, investigations have focused on using ChatGPT for misinformation and hate speech detection \cite{10.1145/3543873.3587368,sallam2023chatgpt}. Other studies have looked at ChatGPT for annotation in a multitude of NLP tasks ranging from sentiment analysis to stance, hate speech, and bot detection~\cite{zhu2023can}. Their findings suggest that while ChatGPT shows promise in generating annotations akin to human judgments, \textit{the accuracy of these annotations remains inconclusive}. 

Moreover, it is important to note that most works are centered on user-generated content from social media. This distinction introduces unique characteristics compared to structured political discourse content~\cite{baldwin2013noisy,deng2021understanding}. Assessing the annotation performance of LLMs in such tasks is not straightforward, due to the diversity of objectives, the subjectivity of natural language, the need for contextual and domain understanding, and the \textit{lack of ground truth or highly-used quality benchmarks}~\cite{kiela2021dynabench,gu2021domain}. The role of benchmark datasets is important as standardized evaluation frameworks, to mitigate entry barriers for new researchers, facilitate comparisons, and foster collaborative work~\cite{harutyunyan2019multitask}. 

In the context of domain-specific benchmarks, the Human ChatGPT Comparison Corpus (HC3) compiles a number of question-and-answer datasets ranging from financial, medical, to psychological domains, and compares ChatGPT’s responses to those of humans~\cite{guo2023close}. However, to the best of the authors' knowledge, no other work with LLMs has specifically targeted textual data within the political science domain for multiple NLP annotation tasks. Yet, the latter is characterized by the use of rhetorical strategies, contextual sensitivity, ideological positions, and formalized structure, distinguishing political discourse from other textual formats \cite{chilton2004analysing}.

$\rightarrow$ \textit{In total, despite their broad applicability, the evaluation of LLMs in annotation tasks for the political discourse domain remains a critical challenge, primarily hinging on the availability of high-quality benchmark datasets.}

\subsection*{Summary of contributions} 
In this paper, we present \mydataset, a meticulously curated, high-quality dataset of 171 campaign speeches tailored for political discourse analysis. The dataset construction employs a two-step "hybrid intelligence" approach: initial automated annotation using ChatGPT, followed by thorough manual human annotation, elevating it to a robust benchmark for critical NLP tasks related to political discourse. To ensure a comprehensive understanding of political discourse, an interdisciplinary team of data journalists, political scientists, and data scientists contribute their expertise. In detail, the \mydataset dataset's main contributions are as follows:
\begin{itemize}
        \item \textbf{\textit{Annotated dataset for political discourse analysis:}} The \mydataset dataset comprises 171 speeches (with 5,279 paragraphs and 717,718 words) from six political party leaders, enriched with metadata, ChatGPT-generated annotations, and human-validated annotations \textit{per paragraph} across six critical NLP tasks: text classification, topic identification, sentiment analysis, polarization and populism detection, and named entity recognition. In total the dataset contains 31,674 annotations, rendering it as a comprehensive benchmark for multidimensional political discourse analysis. The dataset is available at \url{https://doi.org/10.5281/zenodo.13957176}.
        
        \item \textbf{\textit{Human-in-the-loop validation of ChatGPT-based annotations:}} We assess the performance of ChatGPT's annotations across all six NLP tasks, with validation provided by experts (data journalists and political scientists). This approach enables a thorough assessment of ChatGPT's performance, achieving high accuracy in sentiment analysis (93\%) and text classification (89\%), while it is less accurate in more complex tasks (e.g., 61\% in topic classification). %Human annotation further enhanced the dataset's quality by cross-checking ChatGPT annotations and re-annotating instances to capture the interdisciplinary characteristics of the political discourse more accurately.
        
        \item \textbf{\textit{Robustness, reproducibility, and code sharing:}} To develop the \mydataset dataset, we employ a methodology grounded in journalism, political science and data science to analyze the characteristics of political rhetoric\cite{troboukis2024hybridintelligencejournalismfindings}. While our dataset focuses on Greek texts (translated into English), the methodology is adaptable to any language, ensuring the study's overall reproducibility. Moreover, we create an informative online tool with insightful interactive visualizations exploring the \mydataset dataset, useful to journalists, researchers, and the general public \cite{eklogesapp}. Along with the raw data, we have open-sourced supplementary scripts for exploratory data analysis (\url{https://github.com/Datalab-AUTH/AgoraSpeech-EDA}).
\end{itemize}

\begin{figure}[htp]
    \centering
    \includegraphics[width=0.7\textwidth,trim={0 10.5cm 0 9cm},clip]{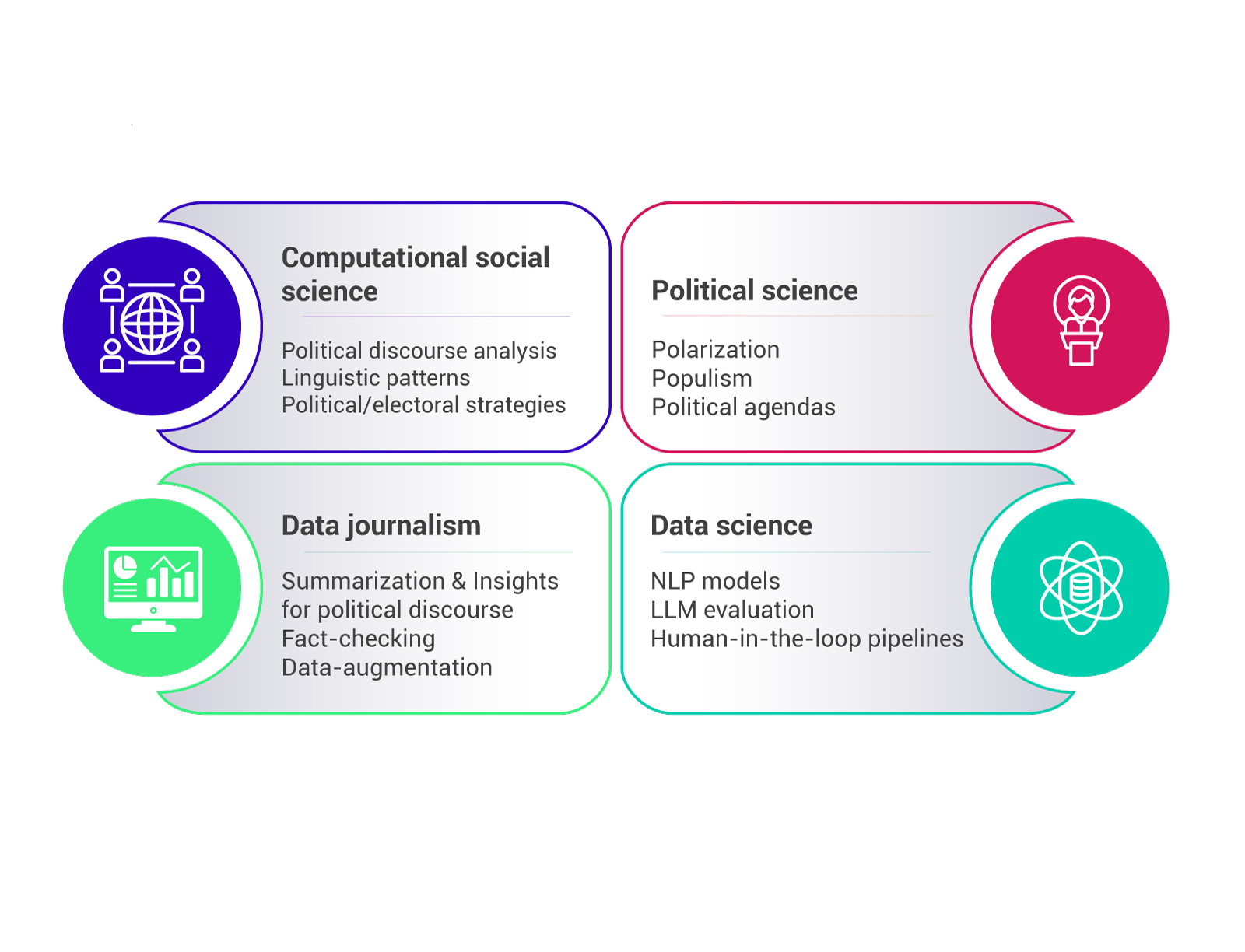}
    \caption{Indicative use cases for the \mydataset dataset.}
    \label{fig:usecases}
\end{figure}

We compile the \mydataset dataset intending to serve multiple-purpose scientific research and to offer valuable applications across several domains (see Figure \ref{fig:usecases}). In \textit{computational social science}, researchers can employ its rich content and methodology to analyze political discourse, gaining insights into language patterns, sentiment, and political strategies. \textit{Data journalists} can utilize the dataset to investigate and report on political speeches and communication strategies. The dataset also provides \textit{political scientists} with a powerful tool for examining global polarization, common political traits, and shared societal dynamics, including trends in populism and polarization. Additionally, it serves as a valuable resource for \textit{data scientists} interested in refining language models and algorithms, contributing to advancements in NLP and model training.

\section*{Methods}\label{methods}
This section provides an overview of the study's methodology, including the data collection process, preprocessing steps, the prompts used for ChatGPT annotations, and the human-in-the-loop evaluation approach.

\subsection*{Study procedure}  % rounds, leaders, multidisciplinary collaboration
The study has strategically synchronized its launching with the onset of the first election period leading up to Greece’s parliamentary elections on May 21, 2023. During this period, we collect and analyze campaign speeches delivered by the leaders of the six political parties that held seats in the parliament. These political figures are (format: Lastname Firstname, Role and Political party):
\begin{itemize}[itemsep=0.05em, labelsep=0.05em]
    \item \textit{Androulakis Nikos}, President of PASOK-Movement for Change
    \item \textit{Koutsoumbas Dimitris}, Secretary General of the Communist Party of Greece %(KKE)
    \item \textit{Mitsotakis Kyriakos}, Prime Minister and President of New Democracy
    \item \textit{Tsipras Alexis}, President of SYRIZA
    \item \textit{Varoufakis Yanis}, Secretary General of MeRA25
    \item \textit{Velopoulos Kyriakos}, President of Greek Solution
\end{itemize}
Following the May 2023 elections, a so-called “one-day parliament” was formed, as no party managed to secure a majority to form a government. Consequently, a second round of elections was scheduled, leading up to the elections on June 25, 2023. The focus of our analysis remains on the same political leaders, excluding Yanis Varoufakis, whose party did not secure a seat in the one-day parliament.

\begin{figure}[htp]
    \centering
    \includegraphics[width=\textwidth]{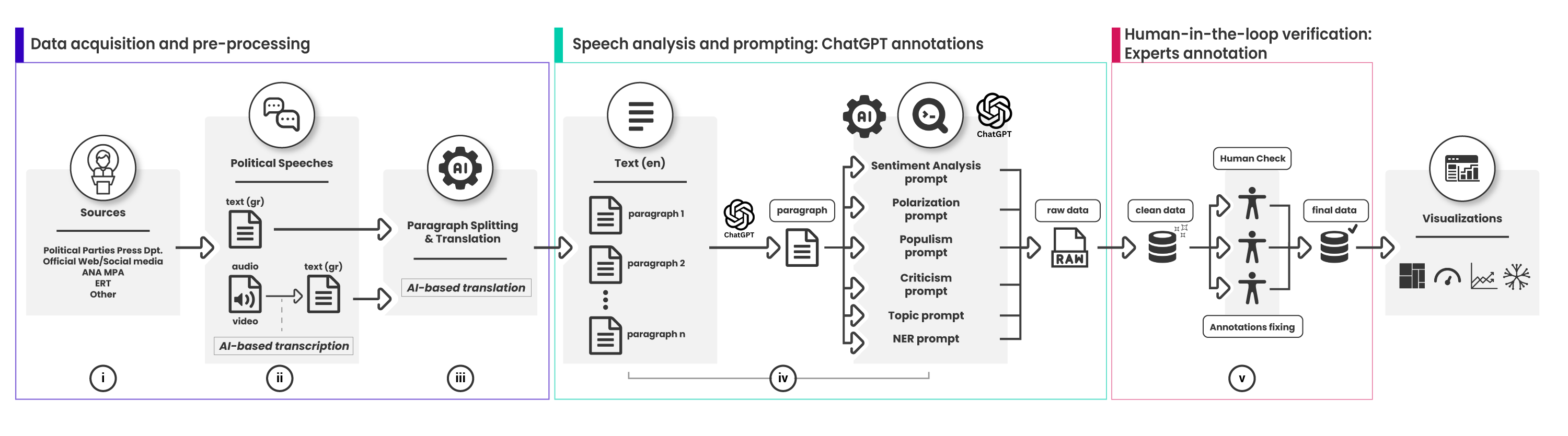}
    \caption{The study procedure for creating the \mydataset dataset, from data acquisition and pre-processing to human-in-the-loop verification.}
    \label{fig:teaser}
\end{figure}

This timeline enables us to build a robust methodology (see Fig. \ref{fig:teaser}) collecting real-time political speeches and developing the insightful \mydataset dataset. This dataset is a collaborative effort between data journalists, a political scientist, and data scientists, leveraging their interdisciplinary expertise for comprehensive political discourse analysis. Data journalists and the political scientist provide real-world insights and analyze the political rhetoric, while data scientists automate parts of the analysis through the use of ChatGPT and advanced political discourse analytics. Indicative outcomes of this collaboration are showcased in an informative online tool\cite{eklogesapp}, demonstrating how \mydataset can serve the wider public.

\subsection*{Data acquisition and pre-processing}  % speech, sources, paragraphs slitting, translation, transcription
To gather the campaign speeches (see Fig. \ref{fig:teaser}(i)), we primarily source them from the press offices of the political parties, followed by their official websites and social media accounts. Additionally, speeches are supplemented with additional information from the Athens-Macedonian News Agency (ANA-MPA), the Hellenic Broadcasting Corporation (ERT), and other local media when they are not fully available through party communication channels.

For this study, any public speech given by the political leaders mentioned earlier, in a public forum with an audience, and during the official pre-election period, is characterized as a campaign speech. However, brief statements or spontaneous talks (i.e., those under 15 minutes) as well as informal interactions with citizens were not included, only full-length campaign speeches that are entirely available are considered for analysis. An important point is that since Kyriakos Velopoulos did not give campaign speeches as defined here, his introductory remarks at various press conferences are analyzed instead, limiting the direct comparison of Velopoulos' speeches with those of other political leaders.

The analysis of the speeches is based on their written transcripts, whenever available (see Fig. \ref{fig:teaser}(ii)). If the provided text is already divided into paragraphs according to topics, we retain this structure. If the speech is not pre-organized into paragraphs or is structured according to oral delivery, our team of data journalists edit and reorganize it by thematic paragraphs. In cases where no written versions of the campaign speeches are available, but audiovisual formats exist, we use Transkriptor, an AI-powered transcription tool (\url{https://transkriptor.com/}), to convert the audio into text. Finally, after collecting and formatting the speeches in Greek, we translate them into English (see Fig. \ref{fig:teaser}(iii)), using a machine translation service, called DeepL (\url{https://www.deepl.com/}), through its API.

\subsection*{Speech analysis and prompting: ChatGPT annotations}
After the required processing, the raw campaign speeches undergo NLP analysis to create the ChatGPT-based annotation benchmark (see Fig. \ref{fig:teaser}(iv)). This paragraph-level analysis focuses on determining whether the leader primarily criticized political opponents or promoted their party's agenda, identifying the main topic or theme, assessing the dominant sentiment, detecting levels of political polarization and populism, and recognizing any named entities. Using an automated way, we prompt ChatGPT (the gpt-3.5-turbo model) to provide paragraph-based outputs for the following NLP tasks:
\begin{itemize}
    \item \textbf{\textit{Text classification}}: for assigning a value of either “criticism" or “political agenda", indicating if the political leader primarily criticized opponents or referred to their party's ideas, opinions, positions, or program proposals.
    \item \textbf{\textit{Topic classification}}: for identifying the most likely theme discussed from a predefined list of 33 specific topics provided to ChatGPT by the data journalists. This list includes the following topics: abstention, accountability, agricultural policy, civil protection, corruption, culture, debt, democracy, economy, education, elections, employment, energy, entrepreneurship, environment, Europe, external affairs, health, human rights, housing, infrastructure, justice, labor, media, migration, national security, pandemic, pensioners, privatization, public sector, social state, transparency, tourism.
    \item \textbf{\textit{Sentiment analysis}}: for assigning a value between -1 and 1, indicating the negativity, neutrality, or positivity of the paragraph, as will be later classified during the human annotation, according to a predefined scale set by the interdisciplinary authors' team.
    \item \textbf{\textit{Polarization detection}}: for assigning a value between 0 to 1, indicating the level of political polarization within the paragraph (no/low, medium, or high), as will be later classified during the human annotation, according to a predefined scale set by the interdisciplinary authors' team.
    \item \textbf{\textit{Populism detection}}: for assigning a value between 0 to 1 to indicate the level of populism within the paragraph (no/low, medium, or high), as will be later classified during the human annotation, according to a predefined scale set by the interdisciplinary authors' team.
    \item \textbf{\textit{Named entities recognition}}: for extracting a list of named entities in the paragraph, categorized as individuals, groups, organizations, political parties, locations, countries, or dates.
\end{itemize}

For accurately annotating the above-presented tasks, ChatGPT prompts are supplemented with additional contextual information, such as the speaker's name and definitions of key terms, to guide the model’s responses and account for specific situations that might influence its output, aiming to bypass ChatGPT's inherent biases. Table \ref{tab:prompts} provides the prompts used for each NLP task. The selection of prompts (including the phrasing of questions and the contextual information) is done with an iterative trial-and-error process that led to the most efficient responses by ChatGPT.

\begin{table}[htp]
\centering
\footnotesize
\caption{The list of prompts (questions and contextual information) used for the ChatGPT annotations.}
\label{tab:prompts}
\begin{tabular}{l|l}
\toprule
\textbf{Task} & \textbf{Prompt content} \\ 
\midrule \midrule
\begin{tabular}{l}
Text \\classification 
\end{tabular}
& 
\begin{tabular}{l}
\textit{Context}: You will read a passage from a political speech of a Greek politician a month before the general elections.\\
\textit{Question}: For the following text respond only with one word, 'criticism' if the speaker criticizes his opponents, or \\ 'political agenda' if the speaker talks about their political agenda or addresses the attendees.
\end{tabular} \\ \hline
\begin{tabular}{l}
Topic \\classification 
\end{tabular}
& 
\begin{tabular}{l}
\textit{Context}: You are reading a passage from a pre-election speech of <Politician's name>, leader of the <Party Name> political \\party.
When you read the term Tempi, this responds to the tragic accident occurred in northern Greece when two trains collided\\ in the village of Tempi, leaving 57 people dead.\\
\textit{Question}: Which of the keywords [<List of Topics>] is closer to the main topic of the following text? Do not provide any \\ explanation, just state the selected keyword. If none of the keywords is relevant, then state the word "other".
\end{tabular} \\ \hline
\begin{tabular}{l}
Sentiment \\ analysis
\end{tabular}
 &  
 \begin{tabular}{l}
\textit{Question:} What is the sentiment of the following text in the range of -1 to 1, where -1 corresponds to negative, 0 to neutral\\ and 1 to positive sentiment? Give me only a single number without any explanations. 
\end{tabular}
\\ \hline
\begin{tabular}{l}
Polarization \\ detection
\end{tabular}
 & 
\begin{tabular}{l}
\textit{Context}: <polarization definition>. \\
\textit{Question}: How polarized is the following text in the range of 0 to 1? Give me only a single number without any explanations. 
\end{tabular}\\ \hline
\begin{tabular}{l}
Populism \\ detection
\end{tabular}
 & 
\begin{tabular}{l}
\textit{Context}: <populism definition>. \\
\textit{Question}: How much populism does the following text have in the range of 0 to 1? Give me only a single number without any\\ explanations.
\end{tabular}
\\ \hline
\begin{tabular}{l}
Named \\ entities \\ recognition
\end{tabular}
& 
\begin{tabular}{l}
\textit{Question}: List the named entities in the following text. List only the entities that belong to only of the types: person, group of\\ people, organization, political party, location, country, date. Write each entry in a new line using the format [named entity, \\type, probability] separated with commas. Do not provide any other explanations or text.
\end{tabular} \\ \bottomrule
\end{tabular}
\end{table}

For instance, when determining whether the text primarily criticized opponents or presented a political agenda, we provide in the prompt the clarification about the type of the given text (which we did not find helpful though for the other tasks). Similarly, when identifying the main topic the prompt is enriched with the political leader's name and status, as well as a description of the accident in Tempi (which we also found it unhelpful for the other tasks). In contrast, names are deliberately omitted when evaluating levels of polarization and populism to avoid ChatGPT's inherent bias. Moreover, to avoid stereotypical interpretations and normative bias, we do not rely on ChatGPT's pre-existing 'knowledge' when assessing polarization and populism, and we provide precise definitions crafted by an expert political scientist~\cite{kiki2023,galanopoulos2023a,galanopoulos2023b}.

\subsection*{Human-in-the-loop verification: Experts annotations}
Through the speech analysis and prompting process, six annotations are generated for each paragraph of every campaign speech, resulting in a comprehensive annotated corpus of 31,674 annotations. These annotations are now undergoing rigorous validation by domain experts (data journalists and a political scientist) (see Fig. \ref{fig:teaser}(v)). This human-in-the-loop approach serves a dual purpose: it refines the output format of the ChatGPT annotations (e.g., for sentiment analysis) and assesses the accuracy and validity of ChatGPT’s annotations. Although the analysis and prompting are based on English translations of the texts, the validation process is conducted using the original Greek texts. This approach allows annotators to identify and correct any inaccuracies. During the human annotation phase, any missing values from the ChatGPT analysis and prompting are supplemented with annotations provided by domain experts. %If disagreements arise among the domain experts, an external expert is consulted to finalize the annotation. 
The domain experts follow a structured set of steps for validating each NLP task in detail:
\begin{itemize}
    \item \textbf{\textit{Text classification}}: Data journalists review each paragraph alongside the ChatGPT-generated classification as either "criticism" or "agenda" and adjust the annotation if needed.%and discuss the validity of the annotation until they reach a consensus. 
    \item \textbf{\textit{Topic classification}}: Like above, data journalists review each paragraph alongside the ChatGPT topic annotation and correct it (or select a more specific topic in the case of "other") if needed.%discuss its validity until they reach a consensus.
    \item \textbf{\textit{Sentiment analysis}}: The data journalists first categorize each paragraph's value-based ChatGPT annotation as "negative" for scores between -1 and -0.34, "neutral" for scores between -0.33 and 0.33, and "positive" for scores between 0.34 and 1. If the ChatGPT sentiment annotation differs from the experts' annotation, the original value is excluded and treated as missing data in the dataset, subsequent analyses, and visualizations. This approach is implemented to avoid numerical human adjustments, ensuring that the sentiment indicator’s average, used to classify the overall campaign speech as negative, neutral, or positive, remains unbiased.
    \item \textbf{\textit{Polarization \& Populism detection}}: The political scientist first categorizes each paragraph's value-based ChatGPT annotation as "no/low" level for scores between 0 and 0.5, "medium" level for scores between 0.51 and 0.8, and "high" level for scores between 0.81 and 1. If the ChatGPT annotation (polarization or populism) level differs from the experts' annotated level, the original value is adjusted, based on the experts' annotation, to 0, 0.6, or 0.9, as appropriate for "no/low" level, "medium" level, and "high" level (of either polarization or populism) classification.

    \textit{Remark: }The majority of excerpts in the "no/low" level category had scores very close to 0. To avoid the potential introduction of extreme values through human intervention within this category, a corrective value of 0 was established for this specific class. Mean scores were used as the corrective values for the remaining categories.
    \item \textbf{\textit{Named entities recognition}}: Many named entities appeared with slight variations, often due to differences in the original text, translation, or ChatGPT’s formatted output. To standardize these variations and ensure consistency, we created a lexicon for all entities that appeared at least twice in the speech corpus. All variations of each entity were then mapped to a single, unified representation.
\end{itemize}

The methods described above provide a robust and reproducible methodology, encompassing the entire process from campaign speech collection to analysis using ChatGPT, complemented by human-in-the-loop validation \cite{troboukis2024hybridintelligencejournalismfindings}. While our use case focuses on Greek political speeches, the methodology is designed to be adaptable to other languages. This rigorous methodology has resulted in the creation of the \mydataset, a high-quality dataset comprising 171 political speeches, divided into 5,279 paragraphs, totaling 31,674 annotations, all systematically evaluated through a human-in-the-loop approach.

\section*{Data Records}
The \mydataset dataset includes information, comprising text, metadata, and speech analysis results, on political speeches delivered during the two pre-election periods of the 2023 Greek National elections, which took place on May 21, 2023, and June 25, 2023.

\begin{table}[htp]
\centering
\footnotesize
\caption{The number of campaign speeches per politician across each election period, along with the percentage per period.}
\label{tab:nofspeeches}
\begin{tabular}{l|c|c||r}
\toprule
\textbf{Politician} &
  \textbf{\begin{tabular}[c]{@{}c@{}}1st elections period\\ (2023-04-22 to 2023-05-21)\end{tabular}} &
  \textbf{\begin{tabular}[c]{@{}c@{}}2nd elections period\\ (2023-05-22 to 2023-06-25)\end{tabular}} &
  \textit{Total} \\
  \midrule \midrule
Androulakis & 16 (15.2\%) & 10 (15.2\%) & 26 \\ \hline
Koutsoumpas & 13 (12.4\%) & 14 (21.2\%) & 27 \\ \hline
Mitsotakis & 26 (24.8\%) & 19 (28.8\%) & 45 \\ \hline
Tsipras & 25 (23.8\%) & 19 (28.8\%) & 44 \\ \hline
Varoufakis & 18 (17.1\%) & - & 18 \\ \hline
Velopoulos & 7 (6.7\%) & 4 (6.1\%) & 11 \\
\midrule \midrule
\multicolumn{1}{r|}{\textit{Total}} & 105 & 66 & \multicolumn{1}{r}{171} \\
\bottomrule
\end{tabular}
\end{table}

In total, the dataset contains 171 political speeches from six politicians. During the first election on May 21, the dataset encompasses 105 speeches, and for the second election on June 25, 66 speeches. Table \ref{tab:nofspeeches} presents the specific number of speeches per politician and election period. It is important to note that Varoufakis is not included in the second pre-election period, as his party did not gain parliamentary representation following the first election.

\subsection*{Data access} 
The \mydataset dataset is stored as a CSV file in Zenodo (\url{https://zenodo.org/}), a general-purpose open repository, and can be accessed online \url{https://doi.org/10.5281/zenodo.13957176}. Each row in the CSV file corresponds to a paragraph from a campaign speech, totaling 5,279 rows, while each column represents a specific type of information related to that paragraph, with 20 columns in total. The dataset provides multiple types of information for each paragraph. First, each paragraph is described by metadata, such as the politician's name, the location of the speech, and other relevant details. Additionally, the paragraph is contextualized by its position within the full campaign speech, with the text available in both Greek and English. Finally, the results of ChatGPT's analysis across six NLP tasks are included, with each task creating a dedicated column. Table \ref{tab:datasetfeatures} offers a detailed overview of the features within the \mydataset dataset.

\begin{table}[htp]
\centering
\footnotesize
\caption{The three information types in the \mydataset dataset: metadata, content, and analysis, accompanied by their features, a short description and values.}
\label{tab:datasetfeatures}
\begin{tabular}{c|l|l|l}
\toprule
\textbf{Information type} &
  \textbf{Feature name} &
  \multicolumn{1}{l|}{\textbf{Description}} &
  \multicolumn{1}{l}{\textbf{Values}} \\ 
  \midrule
\multirow{5}{*}{Metadata} &
  elections &
  The election period of the speech & \begin{tabular}[c]{@{}l@{}}
  "1st Elections 2023-05-21" or \\ "2nd Elections 2023-06-25" \end{tabular}\\ \cline{2-4} &
  speech\_id &
  The ID of the speech & \begin{tabular}[c]{@{}l@{}} string in format: \\ PoliticianName\_YYYY\_MM\_DD\_Location \end{tabular}  \\ \cline{2-4} &
  politician &
  The name of the politician that gave the speech &
  \begin{tabular}[c]{@{}l@{}}"Androulakis", "Koutsoumpas", "Mitsotakis", \\ "Tsipras", "Varoufakis", or "Velopoulos"\end{tabular} \\ \cline{2-4} &
  date (YYYY-MM-DD) &
  The date of the speech &
  string in format: YYYY-MM-DD \\ \cline{2-4} &
  location &
  The location where the speech took place &
  cities in Greece \\ 
  \midrule \midrule
\multirow{3}{*}{Content} &
  paragraph &
  The number of the paragraph in the speech &
  an integer (starting from 1) \\ \cline{2-4} &
  text &
  The text of the paragraph (in english) &
  english text \\ \cline{2-4} &
  text\_el &
  The text of the paragraph (in Greek) &
  greek text \\
  \midrule
  \midrule
\multirow{6}{*}{Analysis} &
  criticism\_or\_agenda &
  \begin{tabular}[c]{@{}l@{}}Whether the text is identified as  \\ "political agenda" or "criticism" \end{tabular} &
  "political agenda" or "criticism" \\ \cline{2-4} &
  topic &
  The main topic of the text &
  string from a predefined list of topics  \\ \cline{2-4} &
  sentiment &
  The detected sentiment of the text &
  a float number between -1 (negative) and 1 (positive) \\ \cline{2-4} &
  polarization &
  The detected level of polarization in the text &
  a float number between 0 (none/low) and 1 (high) \\ \cline{2-4} &
  populism &
  The detected level of populism in the text &
  a float number between 0 (none/low) and 1 (high) \\ \cline{2-4} &
  named entities &
  \begin{tabular}[c]{@{}l@{}}The detected entities (people, locations, \\ organizations, etc.) in the text\end{tabular} &
  string values detected as entities along with metadata \\
  \bottomrule
\end{tabular}
\end{table}

\textit{Remark}: Table \ref{tab:datasetfeatures} presents only 14 features (cf. the dataset contains 20 features). The first 8 features, i.e., metadata and content, are created before any annotations. The remaining 6 features correspond to NLP tasks, each appearing twice in the dataset: once as annotated by ChatGPT and once following verification through the human-in-the-loop process.

\subsection*{Use case: Elections political discourse}
The \mydataset dataset, collected during the 2023 Greek elections across two phases, has already been applied in a compelling real-world case study. The leader-specific and time-sensitive data between election periods offer valuable insights into the evolving political strategies of each leader over time. This analysis resulted in the creation of an informative online tool\cite{eklogesapp} that also functions as an exploratory dashboard for the dataset, allowing users to draw meaningful conclusions from various visualizations. For instance, the analysis disproved early expectations of heightened polarization and populism during the pre-election period, showing only occasional spikes, such as Koutsoumpas' increased polarization during the second election, while Tsipras significantly reduced his populist rhetoric between the two rounds. This real-time analysis provided the public with a clearer understanding of the political landscape, helping voters make more informed decisions. Political scientists and journalists would have gained valuable insights to interpret shifting strategies, allowing for more accurate reporting and deeper analysis of election dynamics. For more detailed outcomes of the use case, we refer the interested reader to the online collection of journalistic articles in the online tool\cite{eklogesapp} and the discussion of the main findings presented in \cite{troboukis2024hybridintelligencejournalismfindings}.

\section*{Technical Validation}
In this section, we present the analyses and experiments conducted to evaluate the technical quality of the \mydataset dataset. We begin with a comparative analysis between ChatGPT and human annotations, followed by assessing the accuracy of ChatGPT's annotations in relation to human annotations. Lastly, we perform an exploratory analysis of the human annotations, which serves as a foundation and motivation for further investigation into the \mydataset dataset.

\subsection*{Data validation: ChatGPT vs. Human annotations}
\begin{figure}[ht!]
    \centering
    \subfloat[Criticism vs. Agenda]{\label{sfig:agendapgt}\includegraphics[width=.49\textwidth]{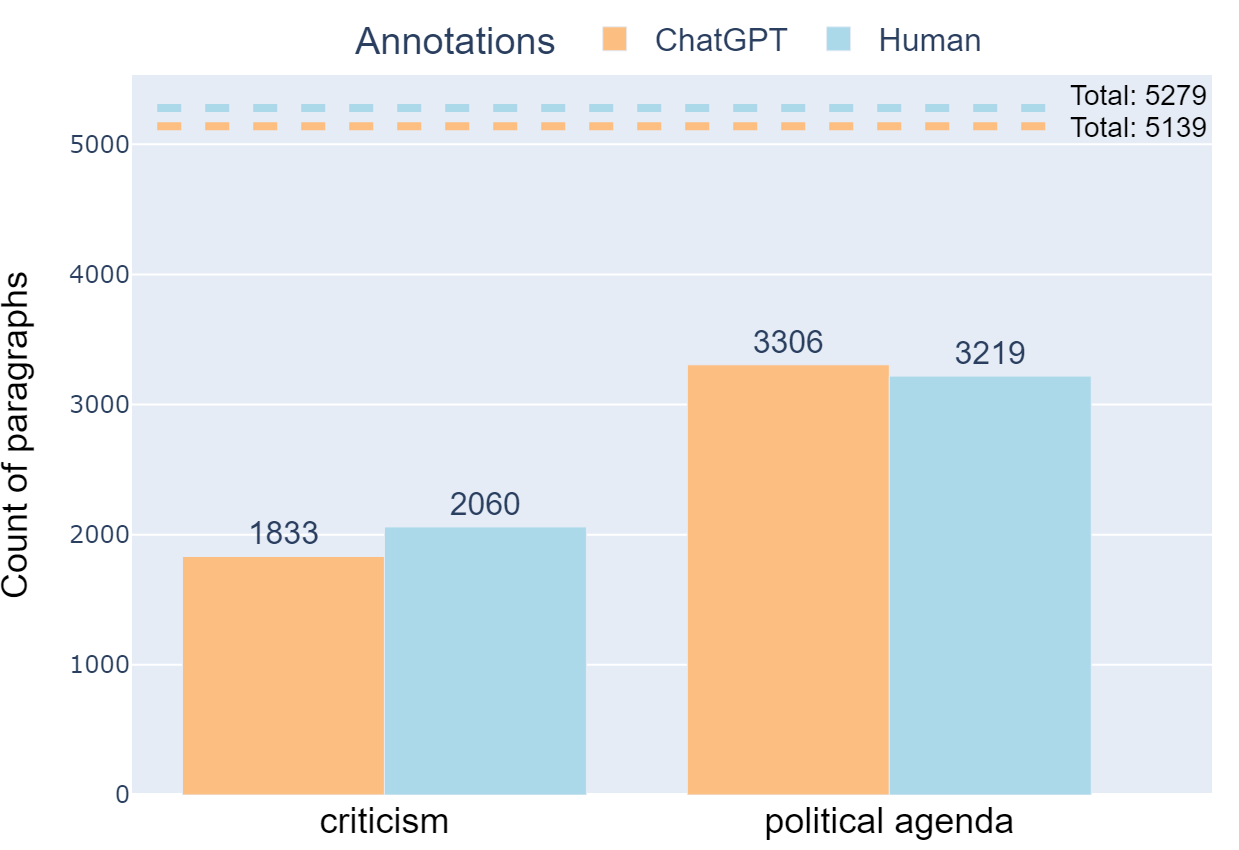}}\hfill
    \subfloat[Topics]{\label{sfig:topicsgpt}\includegraphics[width=.49\textwidth]{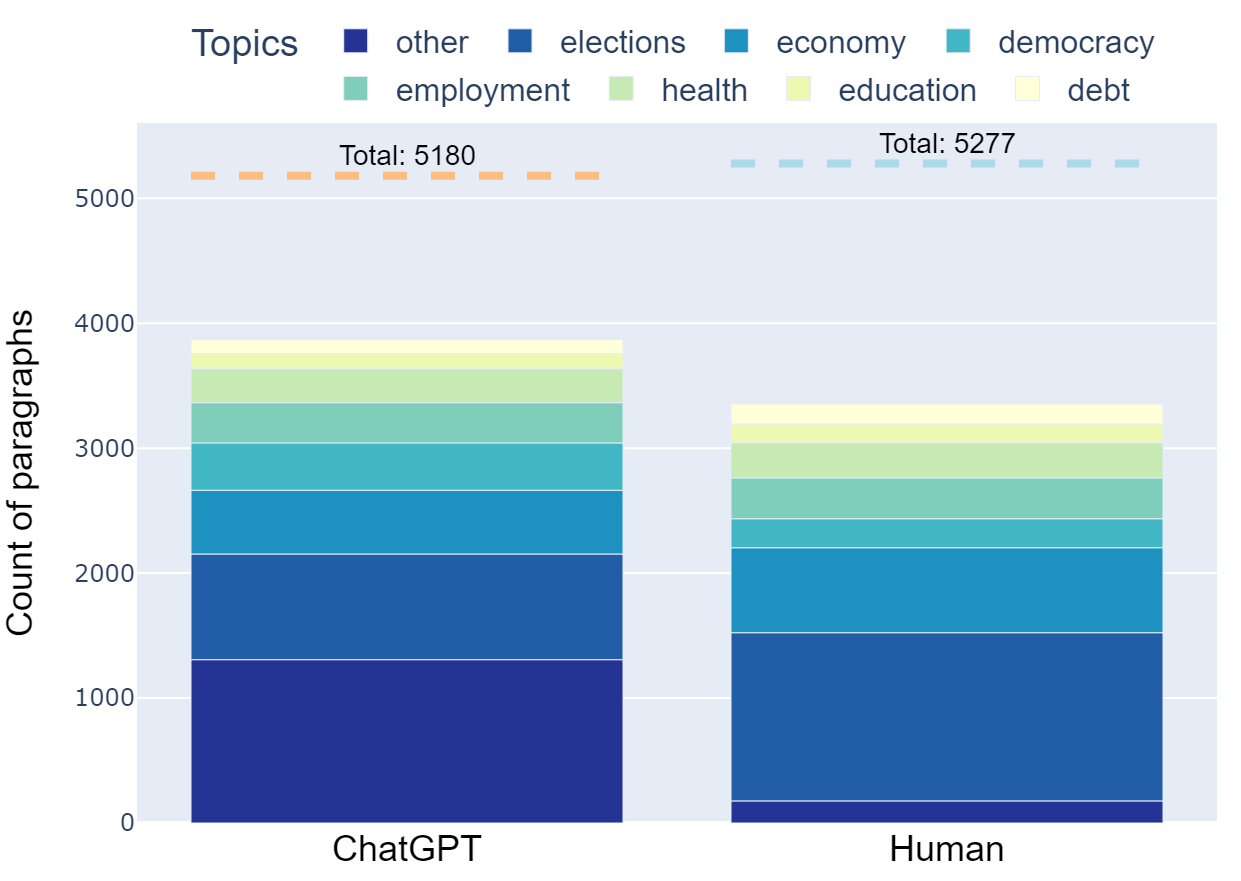}}\\
    \subfloat[Sentiment]{\label{sfig:sentimentgpt}\includegraphics[width=.49\textwidth]{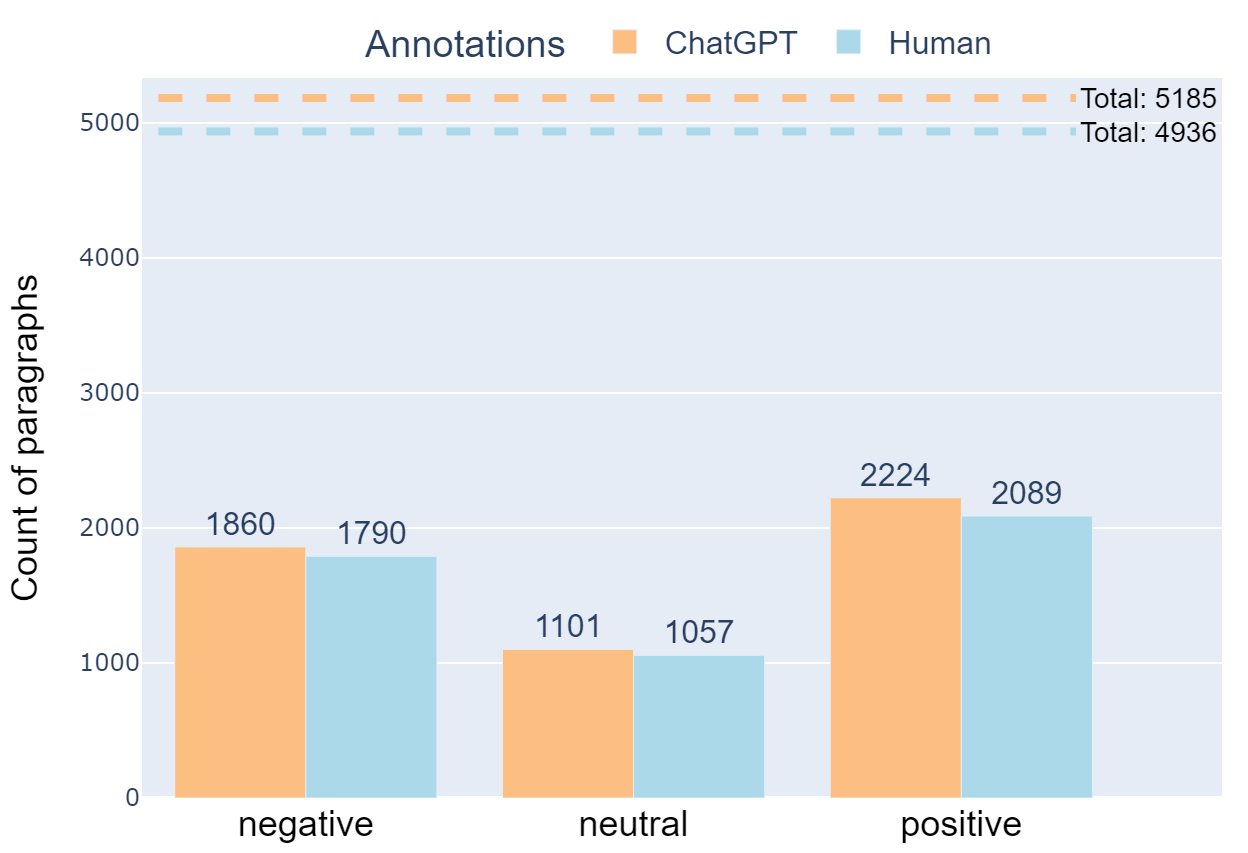}}\hfill
    \subfloat[Polarization]{\label{sfig:polarizationgpt}\includegraphics[width=.49\textwidth]{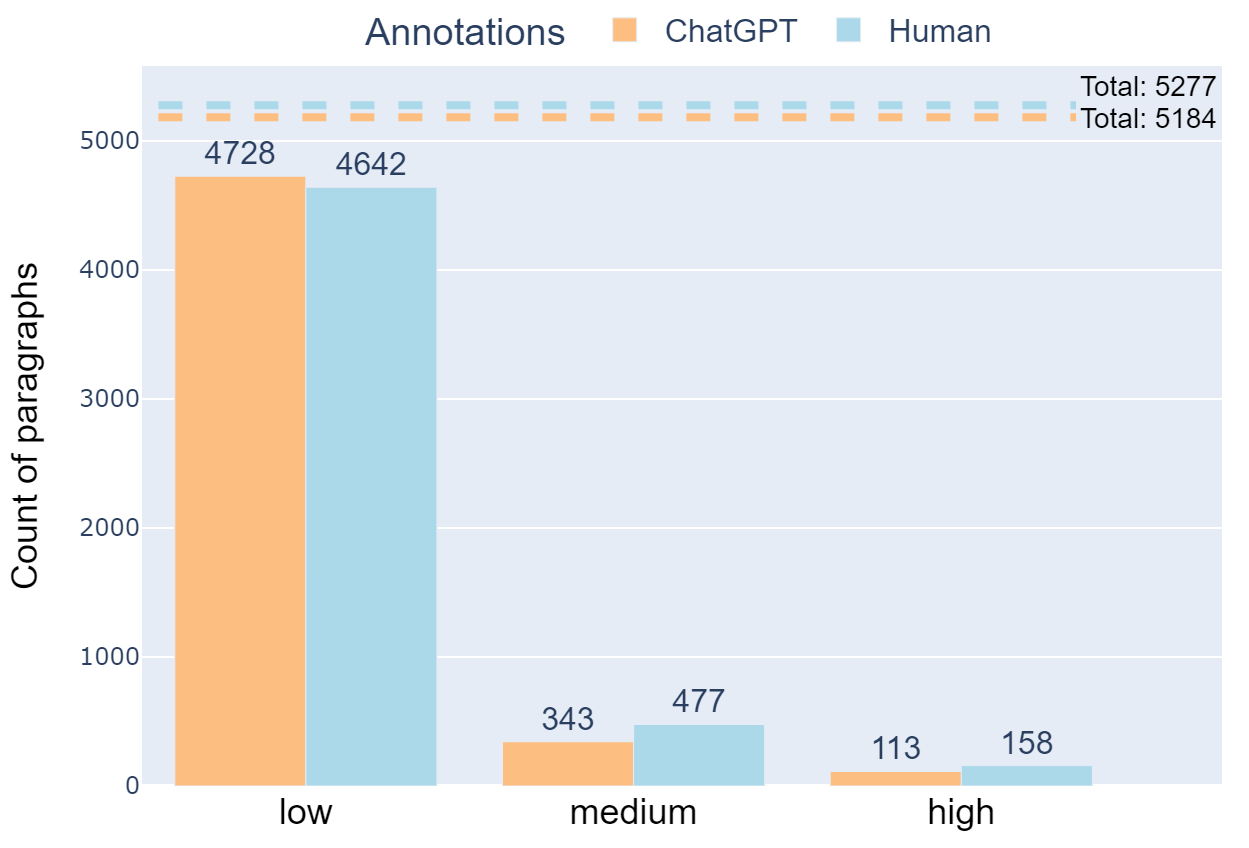}}\\
    \subfloat[Populism]{\label{sfig:populismgpt}\includegraphics[width=.49\textwidth]{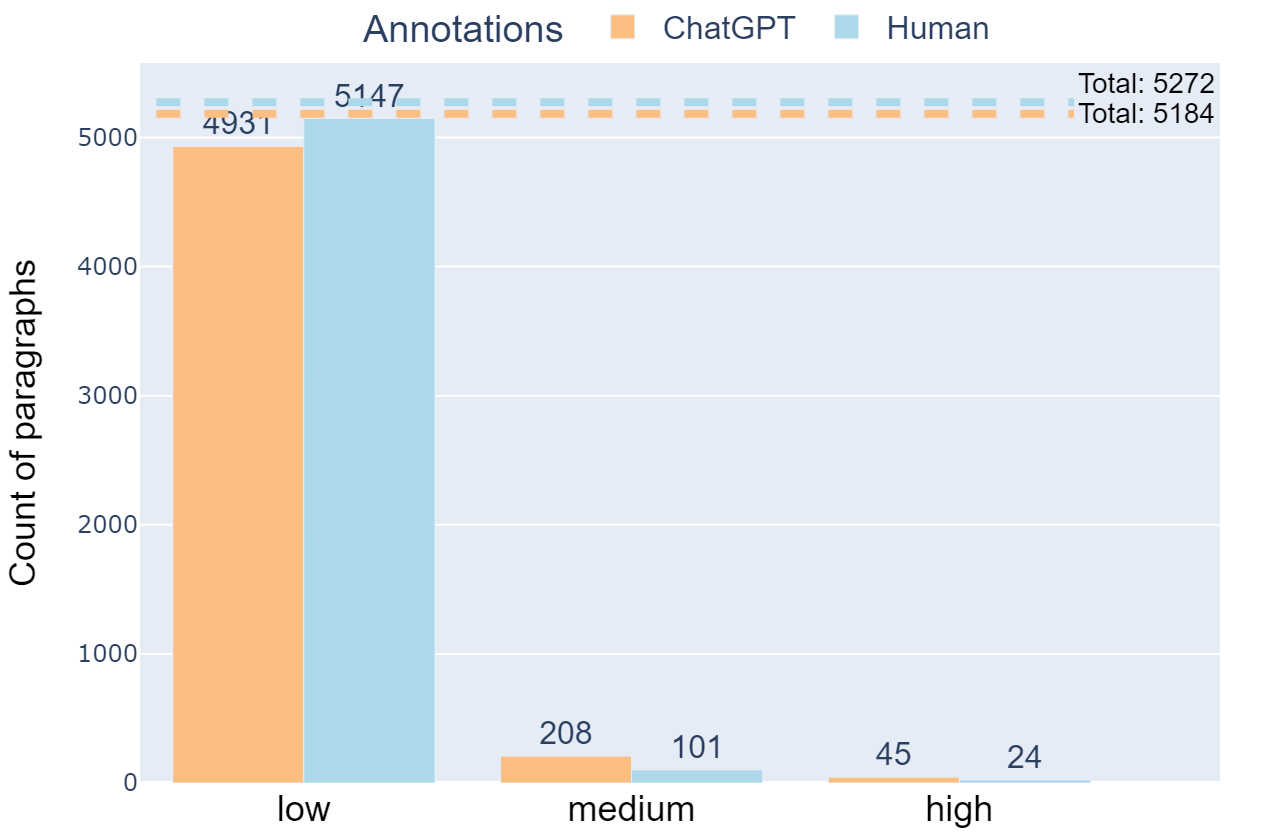}}\\
    \caption{The distribution of values across the NLP tasks, as assigned by ChatGPT annotations (orange/left bars) and after the human-in-the-loop annotation process (blue/right bars), for the overall dataset.}
    \label{fig:gptvshuman}
\end{figure}

As outlined in the \textit{Methods} section, the annotation of the NLP tasks follows a two-step process. Initially, ChatGPT performs the annotation using tailored prompts enriched with contextual information (see Table \ref{tab:prompts}), and in the second step, human experts cross-check and re-annotate all instances across the NLP tasks. To analyze the results of the two-step annotation process, Figure \ref{fig:gptvshuman} presents the distribution of values across the NLP tasks, along with the total non-missing annotations, comparing the ChatGPT and human annotations. For the analysis and visualization, continuous features (sentiment, polarization, and populism) are transformed into categorical variables, as described in the \textit{Methods} section. For the named entity recognition task, since the values span a large range and the annotation by human experts follows a different process, its results are excluded from the following figures and are discussed later in the paper (see, e.g., Figure~\ref{sfig:stackedhorizontal}).

\myitem{Text classification:} Starting from the text classification into "criticism" or "political agenda" (see Figure \ref{sfig:agendapgt}), the majority of paragraphs are labeled as "political agenda" in both annotation steps. However, ChatGPT assigns more paragraphs to "political agenda" compared to human experts. This suggests that human experts re-annotate some instances initially classified by ChatGPT as "political agenda" into "criticism", highlighting a potential weakness in ChatGPT's ability to identify a critical tone or related terms in certain political speeches. Additionally, human experts annotate the 140 paragraphs that were missed by ChatGPT in the initial step of annotation. 

\myitem{Topic classification:} Moving to the topic classification into one of the 33 predefined topics, Figure \ref{sfig:topicsgpt} depicts the most discussed topics, which follow a similar distribution between ChatGPT and human experts, except for the "Other" category. This suggests that human experts re-annotate some instances initially classified by ChatGPT as "Other" into more specific topics, highlighting a potential limitation in ChatGPT's ability to categorize text into narrower topics. Additionally, human experts annotate 97 paragraphs that were missed by ChatGPT in the initial step (due to technical or inaccuracy reasons), though annotations for 2 paragraphs remain missing. 

\myitem{Sentiment analysis:} In terms of sentiment analysis and classification into "negative," "neutral," and "positive" (see Figure \ref{sfig:sentimentgpt}), the distribution of values is similar between ChatGPT and human annotations, suggesting that ChatGPT is efficient for sentiment analysis in political discourse. The number of annotated paragraphs is lower after the human cross-check since human annotators left empty values when they observed an inaccuracy by ChatGPT to avoid bias in further analysis. This results in 343 missing values in the final annotations for the sentiment analysis task.

\myitem{Polarization and Populism detection:} Differences in polarization identification as "no/low", "medium", and "high" are evident (see Figure \ref{sfig:polarizationgpt}), with the majority of paragraphs identified as "no/low" in both annotation steps. However, ChatGPT assigns more paragraphs to the "no/low" polarization level and fewer to the "medium" and "high" levels compared to human experts. This suggests that human experts re-annotate some instances initially classified as "no/low" by ChatGPT into "medium" and "high", highlighting a potential weakness in ChatGPT's ability to detect polarized tones in certain political speeches. Additionally, human experts annotate 93 paragraphs missed by ChatGPT in the initial step, though 2 paragraphs remain missing. Similarly, in the identification of populism levels (see Figure \ref{sfig:populismgpt}), the majority of paragraphs are classified as "no/low" in both annotation steps. However, unlike polarization, ChatGPT assigns fewer paragraphs to the "no/low" populism level and more to the "medium" and "high" levels compared to human experts. This leads human experts to re-annotate some instances initially classified by ChatGPT as "medium" and "high" into "no/low," suggesting a tendency for ChatGPT to identify populist rhetoric where it may not exist at those levels. Additionally, human experts annotate 88 paragraphs missed by ChatGPT in the initial step, though 7 paragraphs remain missing.

\begin{graycomment}
    Human experts re-annotated instances in text classification from "political agenda" to "criticism", in topic classification from the "Other" category to more specific categories, in polarization detection from "no/low" to "medium" or "high", and in populism detection from "medium" or "high" to "no/low".
\end{graycomment}

In summary, the exhaustive human annotation process (31,674 annotations, in total, for all paragraphs and NLP tasks) aimed to address potential limitations in ChatGPT annotations, as those described above, since the efficiency of LLM-based annotations for political discourse has not been tested before. The findings from the comparative analysis in Figure \ref{fig:gptvshuman} show that human annotation led to several corrections and differences for the final dataset. To obtain more insights into the efficiency of ChatGPT annotations, we perform a quantitative analysis, where we evaluate the accuracy of ChatGPT with respect to human annotations. Specifically, we calculate the accuracy as the fraction of paragraphs where ChatGPT's annotation remains unchanged by human reviewers, relative to the total number of paragraphs, as follows:
\begin{equation}
    \text{accuracy (task X)} = \frac{\textit{number of paragraphs where ChatGPT and human annotations for task X match each other}}{\textit{total number of paragraphs}}
\end{equation}
    
The accuracy results, expressed as percentages, are presented in Table \ref{tab:chatgptaccuracy}, along with details about value ranges, classes, and the frequency of prevalent values for each NLP task. This frequency reflects the accuracy of a baseline "dummy" model, which would always select the most common among the final human annotations in each NLP task, providing a reference point for evaluating ChatGPT's performance. In an ideal scenario, where ChatGPT's annotations perfectly match those of human annotators, the accuracy would be 100\%.
\begin{table}[htp]
    \centering
    \footnotesize
    \caption{Accuracy of ChatGPT annotations across the NLP tasks, along with details on the possible values (i.e., classes of the classification NLP task) and most common values.}
    \label{tab:chatgptaccuracy}
    \begin{tabular}{l|l|c|c}
    \toprule
    \textbf{NLP task} & \textbf{Values ranges and classes} & \multicolumn{1}{c|}{\textbf{Prevalent value}} & \multicolumn{1}{c}{\textbf{Accuracy}} \\
    \midrule \midrule
    Text classification & "criticism", "political agenda" & political agenda (61\%) & 89\% \\ \hline
    Topic classification & predefined list of 33 topics & elections (25\%) & 61\% \\ \hline
    Sentiment analysis & "negative", "neutral", "positive" & positive (40\%) & 93\% \\ \hline
    Polarization detection & "no/low", "medium", "high" & no/low (88\%) & 88\% \\ \hline
    Populism detection & "no/low", "medium", "high" & no/low (97\%) & 93\% \\
    \bottomrule
    \end{tabular}
\end{table}

From Table \ref{tab:chatgptaccuracy}, it is clear that ChatGPT demonstrates strong performance in text classification, achieving an impressive 89\% accuracy in distinguishing between politicians critiquing their opponents and presenting their political agenda. However, in the more complex task of topic classification, where ChatGPT assigns a paragraph to one of 33 predefined topics, accuracy drops to 61\%. While this may appear relatively low, it is important to consider the inherent difficulty of multiclass classification with 33 categories, ChatGPT still outperforms a baseline "dummy" model by almost threefold. Furthermore, in sentiment analysis, ChatGPT excels with a high accuracy of 93\%, the highest among all tasks, demonstrating its reliability in identifying sentiment within political discourse. This aligns with recent research findings on ChatGPT’s effectiveness in sentiment analysis \cite{zhu2023can, bang2023multitask}. For polarization detection, ChatGPT also shows high accuracy (88\%), though this performance is comparable to a "dummy" classifier, which would label most paragraphs as “no/low” due to the high prevalence of that category. A similar trend is observed in populism detection, where ChatGPT achieves a high accuracy of 93\%, but a "dummy" classifier would be even more accurate (97\%) given that most paragraphs fall into the "no/low" populism category. These findings suggest that ChatGPT may not be ideal for unsupervised detection of nuanced political speech concepts like polarization or populism. In fact, ChatGPT tends to overestimate the levels of these attributes, likely offering results aligned with broader interpretations of these concepts, as opposed to the precise definitions used in political science.

\begin{graycomment}
    ChatGPT demonstrates impressive performance with a 93\% accuracy in sentiment analysis, high performance with 89\% in text classification and 88\% in polarization detection, and a competitive 61\% in topic classification.
\end{graycomment}

\subsection*{Detailed observations of human annotations}
After the human-in-the-loop annotation process, the \mydataset dataset contains a total of 31,674 high-quality human annotations across 5,279 paragraphs for six NLP tasks. These annotations can offer valuable insights for interdisciplinary political discourse analysis, not only for the dataset as a whole but also when broken down per politician. In this context, Table \ref{tab:basic-stats} provides descriptive statistics for all politicians included in the dataset, with a detailed breakdown for each. Moreover, the analysis, discussion and visualizations presented below, further highlight the richness and quality of the human annotations, and how they can be used to reveal key patterns in political discourse (here, for the Greek national elections in 2023).

\begin{table}[htp] 
    \centering
    \caption{Descriptive statistics for all politicians and individually per politician.}
    \label{tab:basic-stats}
    \footnotesize
    \begin{tabular}{l|c|c|c|c|c|c|c}
        \toprule
        \textbf{Statistic} & \textbf{Overall} & \textbf{Androulakis} & \textbf{Koutsoumpas} & \textbf{Mitsotakis} & \textbf{Tsipras} & \textbf{Varoufakis} & \textbf{Velopoulos} \\
        \midrule \midrule
        Total number of speeches & 171 & 26 & 27 & 45 & 44 & 18 & 11 \\ \hline
        Total number of paragraphs & 5,279 & 883 & 707 & 1,385 & 1,601 & 525 & 170 \\ \hline
        Total number of words & 717,718 & 132,620 & 98,807 & 147,709 & 185,139 & 124,716 & 28,727 \\ \hline
        % Avg paragraphs per speech & 30.87 & 33.96 & 26.19 & 30.78 & 36.39 & 29.17 & 16.18 \\ \hline
        Avg paragraphs per speech & 31 & 34 & 26 & 31 & 36 & 29 & 16 \\ \hline
        Avg words per speech & 4,197 & 5,101 & 3,660 & 3,282 & 4,208 & 6,929 & 2,612 \\ \hline
        Avg words per paragraph & 136 & 150 & 140 & 107 & 116 & 238 & 161 \\ \hline
        Agenda percentage & 61\% & 60\% & 52\% & 85\% & 54\% & 51\% & 29\% \\ \hline
        Criticism percentage & 39\% & 40\% & 48\% & 15\% & 46\% & 49\% & 71\% \\ \hline
        Sum of unique topics of all speeches & 33 & 32 & 31 & 31 & 31 & 28 & 27 \\ \hline
        % Avg of unique topics per speech & 12.53 & 15.58 & 9.85 & 12.47 & 13.89 & 11.00 & 9.18 \\  \hline      
        Avg of unique topics per speech & 13 & 16 & 10 & 12 & 14 & 11 & 9 \\  \hline      
        Avg sentiment score & 0.03 & 0.00 & -0.25 & 0.46 & -0.05 & -0.27 & -0.44 \\ \hline
        % Avg sentiment score per speech & 0.01 & -0.01 & -0.25 & 0.49 & -0.06 & -0.29 & -0.44 \\ \hline
        Avg polarization score & 0.16 & 0.17 & 0.26 & 0.07 & 0.16 & 0.18 & 0.20 \\ \hline
        % Avg polarization score per speech & 0.16 & 0.16 & 0.26 & 0.07 & 0.16 & 0.18 & 0.21 \\ \hline
        Avg populism score & 0.07 & 0.09 & 0.11 & 0.02 & 0.09 & 0.09 & 0.05 \\ \hline
        % Avg populism score per speech & 0.07 & 0.09 & 0.12 & 0.02 & 0.09 & 0.09 & 0.05 \\ \hline
        Sum of unique entities of all speeches & 7,763 & 1,605 & 1,531 & 2,050 & 2,597 & 1,911 & 600 \\ \hline
        % Avg of unique entities per speech & 122.61 & 137.92 & 104.96 & 99.96 & 129.84 & 192.67 & 78.91 \\ \hline
        Avg of unique entities per speech & 123 & 138 & 105 & 100 & 130 & 193 & 79 \\ 
        % Avg of unique entities per paragraph & 5.66 & 6.51 & 5.97 & 4.51 & 4.99 & 8.53 & 6.74 \\
        % Avg of unique entities per paragraph & 5.7 & 6.5 & 6.0 & 4.5 & 5.0 & 8.5 & 6.7 \\
        \bottomrule
    \end{tabular}
\end{table}

\subsubsection*{Do politicians focus on presenting their agenda or on criticizing their opponents? (text classification)}
\begin{figure}[htp]
    \centering
    \includegraphics[width=0.8\textwidth]{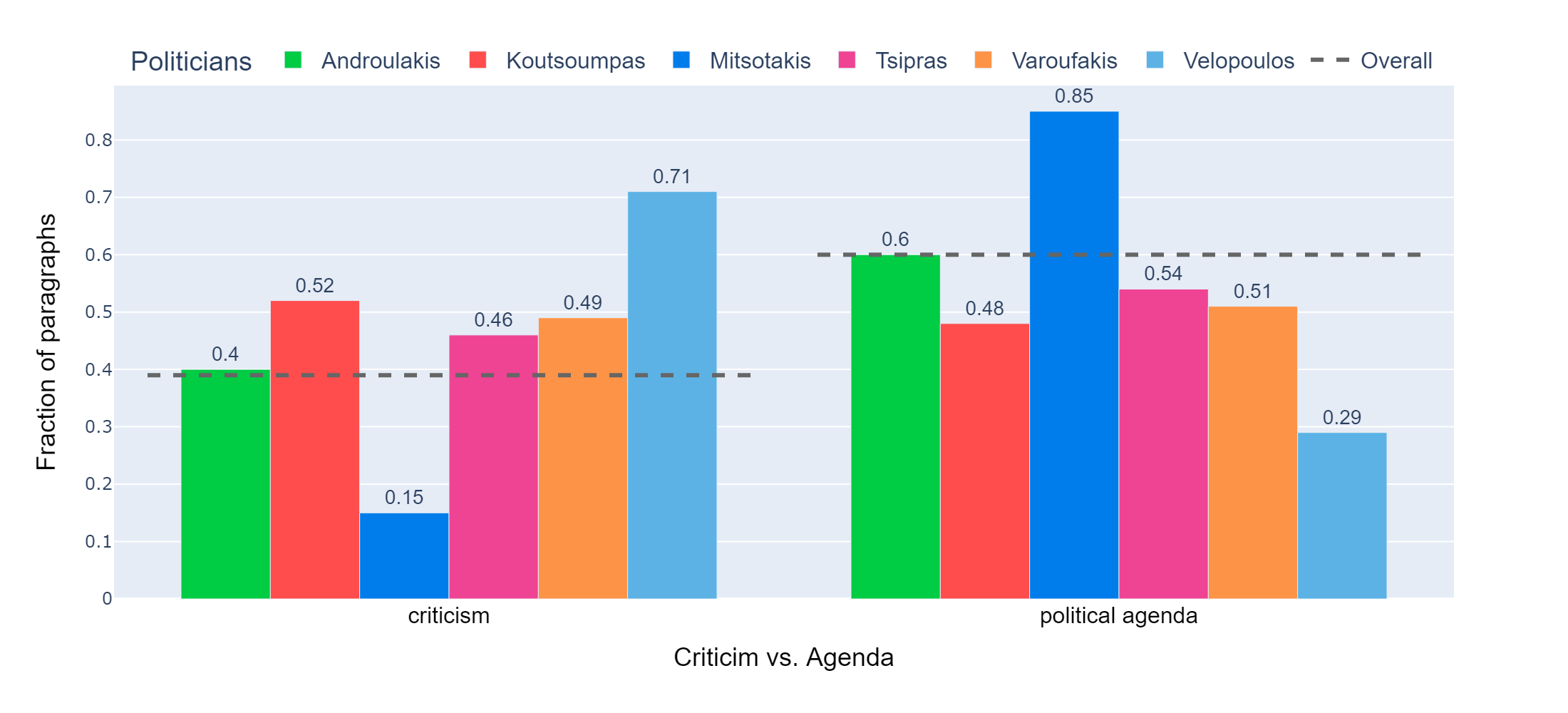}
    \caption{Fraction of the political speeches devoted to "criticism" vs. " political agenda" for each politician in the dataset.}
    \label{fig:criticism_agenda}
\end{figure}

For comparing the narrative strategies of six politicians, Figure \ref{fig:criticism_agenda} presents the fraction of paragraphs devoted to either criticism or political agenda. Mitsotakis stands out with 85\% of his narrative dedicated to policy discussions and only 15\% to criticism, suggesting a strong emphasis on outlining governance plans as prime minister. In contrast, Velopoulos is the most criticism-focused, with 71\% of his speech dedicated to critique and only 29\% focused on his own agenda. The other politicians show a more balanced approach. Androulakis devotes 40\% to criticism and 60\% to agenda, while Tsipras follows closely with 46\% on criticism and 54\% on agenda. Koutsoumpas leans slightly toward criticism (52\%), and Varoufakis is nearly balanced, with 49\% on criticism and 51\% on agenda. 

%Overall, most politicians demonstrate a mix of both criticism and agenda in their discourse, with Mitsotakis and Velopoulos positioned at opposite extremes. The sharp contrast between these two is particularly striking: Mitsotakis is heavily focused on promoting his political agenda, while Velopoulos prioritizes criticism. This difference highlights their distinct political strategies. In contrast, the more balanced approaches seen in Androulakis, Tsipras, Koutsoumpas, and Varoufakis suggest a more versatile approach, blending critique with policy proposals.

\begin{graycomment}
    Most politicians strike a balance between presenting their agenda and criticizing opponents, though Mitsotakis strongly focuses on his agenda, while Velopoulos prioritizes criticism, reflecting differing political strategies.
\end{graycomment}

\subsubsection*{In what topics do politicians focus on? (topic classification)}
\begin{figure}[htp]
    \centering
    \includegraphics[width=0.8\textwidth]{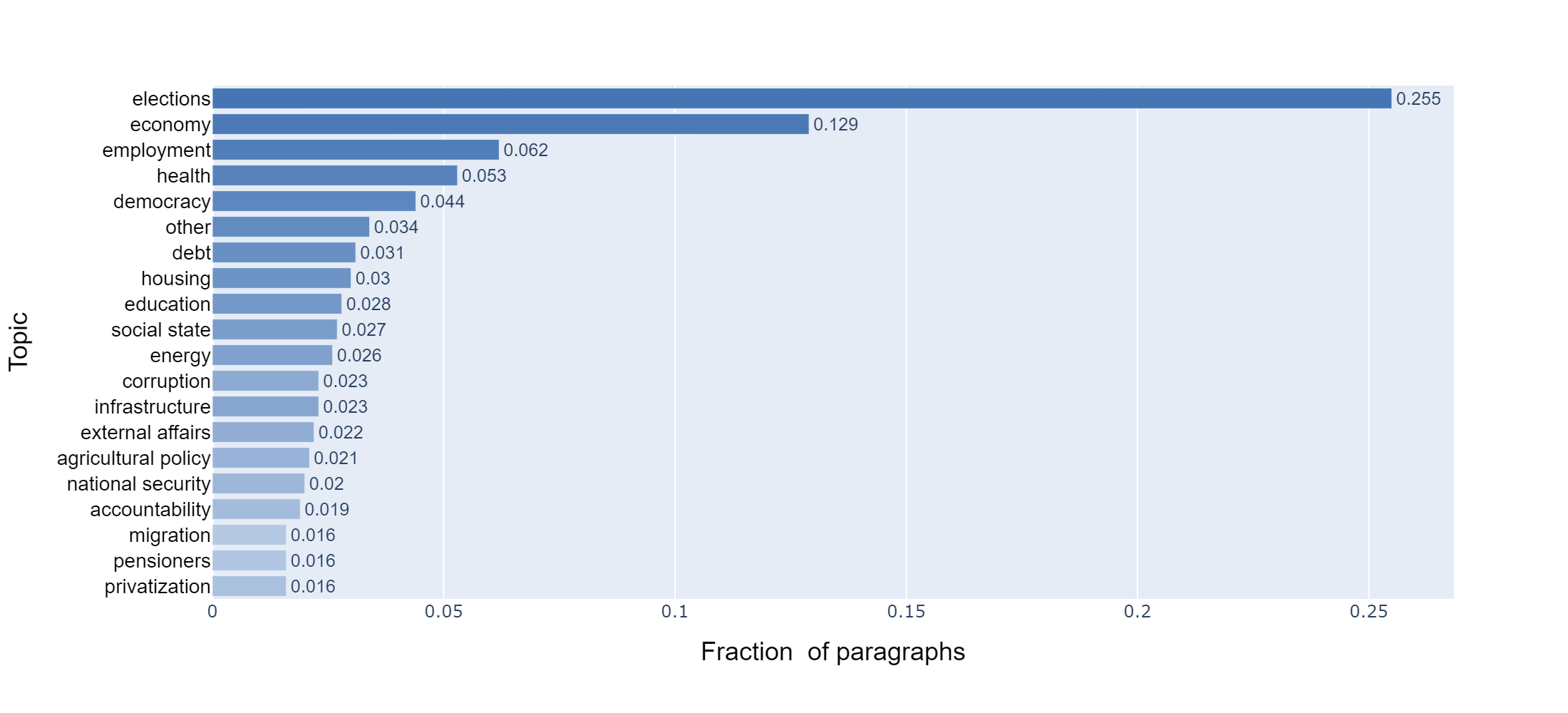}
    \caption{The distribution of the 20 most frequent topics across the campaign speeches of the dataset overall.}
    \label{fig:topicshorizontal}
\end{figure}

For identifying the dominant topic in politicians' campaign speeches, Figure \ref{fig:topicshorizontal} presents the distribution of paragraphs across various topics. Elections dominate the discussion at 25.5\%, followed by the economy at 12.9\%. Employment (6.2\%), health (5.3\%), and democracy (4.4\%) also receive significant attention. In contrast, topics like corruption and national security, among others receive only around 2\%, with migration and privatization discussed even less at 1.6\%, indicating a focus on obvious topics while sidelining other significant issues.

\begin{figure}[htp]
    \centering
    \includegraphics[width=0.8\textwidth]{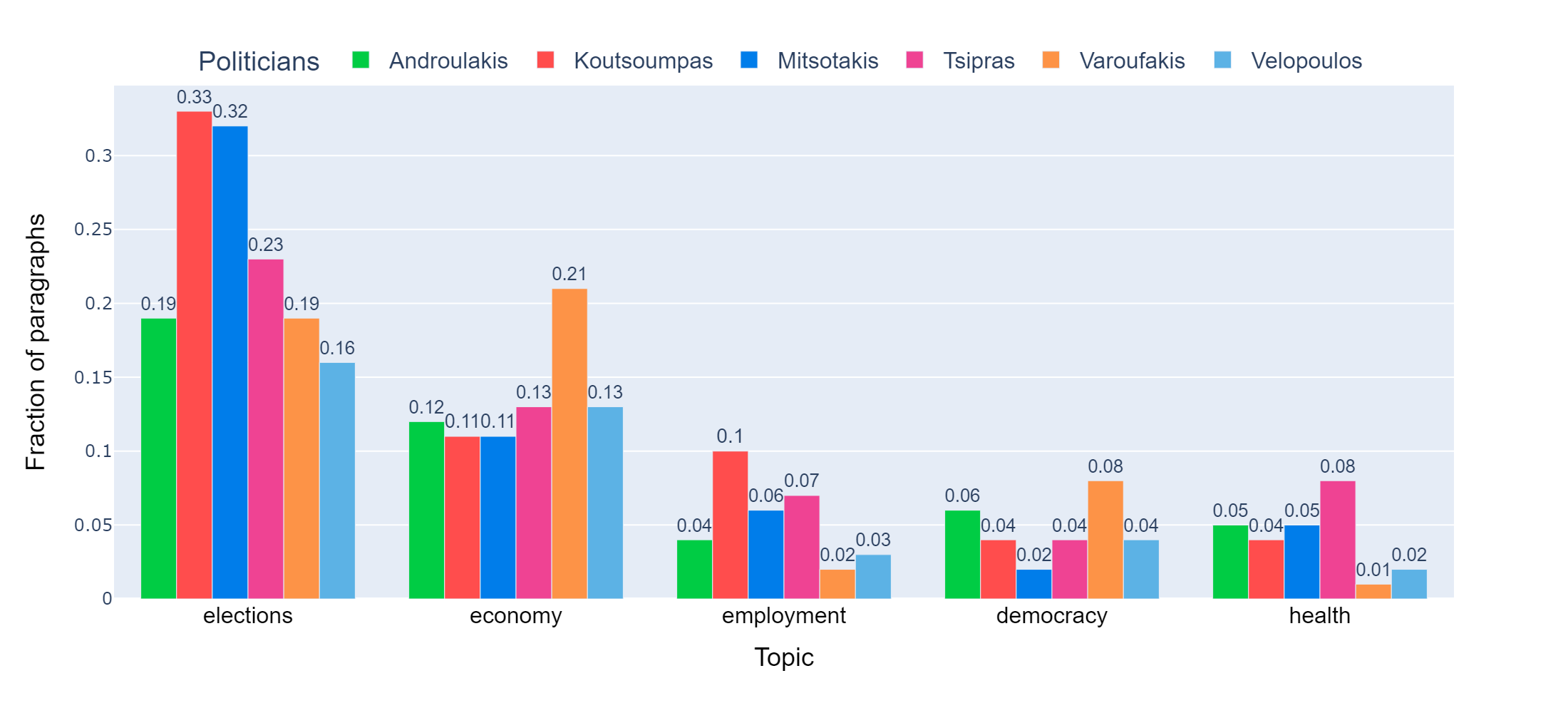}
    \caption{The distribution of the 5 most frequent topics across the campaign speeches per politician.}
    \label{fig:topicsgrouped}
\end{figure}

Figure \ref{fig:topicsgrouped} compares the fraction of paragraphs devoted to the 5 most frequent topics -elections, economy, employment, democracy, and health- by the six politicians. Koutsoumpas and Mitsotakis prioritize elections, with 33\% and 32\% of their discourse, followed by Varoufakis (23\%) and Tsipras (19\%). Varoufakis leads on the economy (21\%), while Mitsotakis, Tsipras, and Velopoulos focus on around 13\%. Koutsoumpas stands out for employment (10\%), with Varoufakis and Tsipras following, while Velopoulos shows the least attention (2\%). On democracy and health, Varoufakis and Tsipras are more engaged (8\%) respectively, while Mitsotakis and Varoufakis trail.

Overall, Koutsoumpas, Mitsotakis, and Tsipras place greater emphasis on widely discussed topics such as elections and the economy, focusing on the most timely and relevant issues. In contrast, Androulakis shows a more balanced approach, covering both the prominent topics and less-discussed but still important issues. Varoufakis, and especially Velopoulos, shift their attention toward less discussed topics, despite giving some focus to democracy and elections, respectively. These patterns underscore the differing political priorities and communication strategies among the politicians.

\begin{graycomment}
    Politicians' speeches focus mainly on elections and the economy, with Koutsoumpas, Mitsotakis, and Tsipras emphasizing these topics, while Androulakis balances prominent and less-discussed issues, and Varoufakis and Velopoulos shift attention to less common topics.
\end{graycomment}

\subsubsection*{What is the prevalent sentiment in politicians' speeches? (sentiment analysis)}
\begin{figure}[htp]
    \centering
    \includegraphics[width=0.8\textwidth]{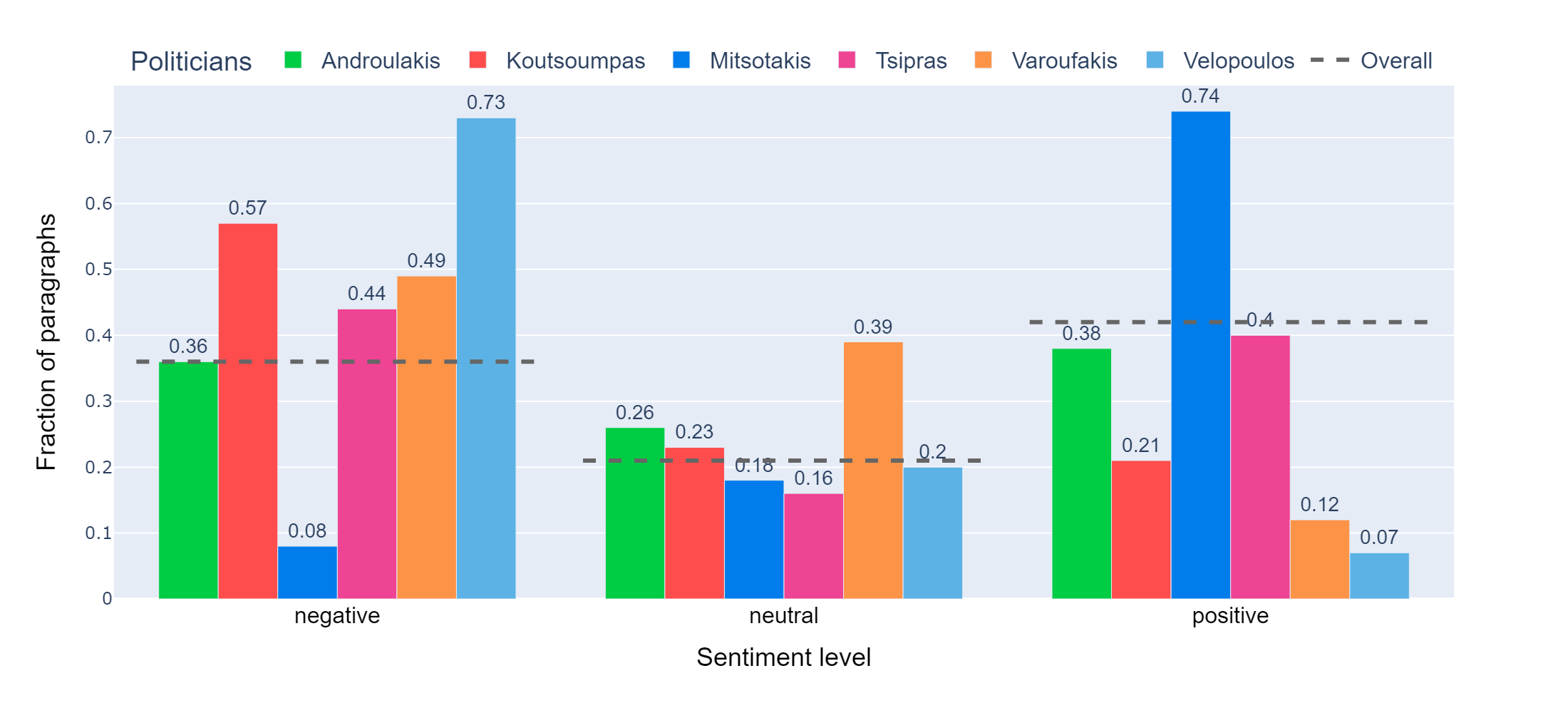}
    \caption{The distribution of sentiment in the campaign speeches within the dataset, broken down by each politician.}
    \label{fig:sentiment}
\end{figure}

For revealing distinct differences in the sentiment of campaign speeches, Figure \ref{fig:sentiment} presents the sentiment distribution among the paragraphs of the six politicians. Mitsotakis stands out with 74\% of his speech having a positive tone, aligning with his prime minister position, and in sharp contrast to Velopoulos, whose discourse has 72\% negative sentiment. Koutsoumpas also leans toward negative sentiments (57\%). Varoufakis uses mostly neutral sentiment (39\%), while Tsipras has the second-highest positive sentiment at 40\%. Androulakis demonstrates a more even distribution of sentiment, with 36\% negative, 26\% neutral, and 38\% positive, indicating a relatively balanced approach compared to his peers.

It is noteworthy the sharp contrast between Mitsotakis and Velopoulos, who represent opposite ends of the sentiment spectrum. Mitsotakis exhibits a positive sentiment level, highlighting his leadership, while Velopoulos adopts a negative tone, aligning with his oppositional role and his focus on criticism over agenda. Varoufakis stands out for his neutral tone, reflecting a more balanced, policy-driven approach, while Androulakis presents a moderate and balanced sentiment. These patterns highlight how each politician’s communication style reflects their political stance and strategies.

\begin{graycomment}
    The prevalent sentiment in politicians' speeches varies widely, with Mitsotakis leaning heavily toward positivity, Velopoulos toward negativity, while Varoufakis and Androulakis adopt more neutral and balanced tones.
\end{graycomment}

\subsubsection*{To what extent are politicians' speeches polarized? (polarization detection)}
\begin{figure}[htp]
    \centering
    \subfloat[Polarization]{\label{sfig:polarization}\includegraphics[width=.49\textwidth]{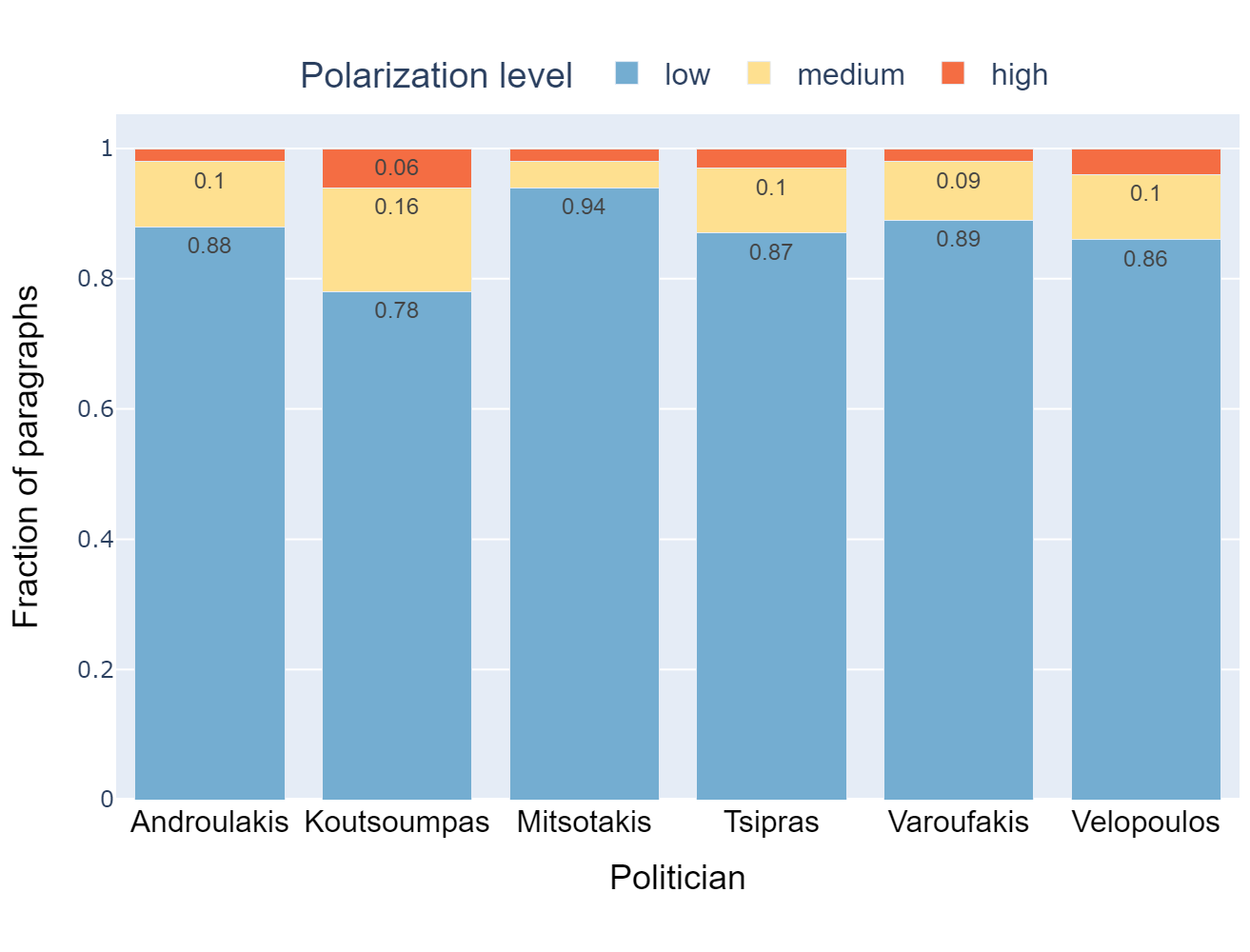}}\hfill
    \subfloat[Populism]{\label{sfig:populism}\includegraphics[width=.49\textwidth]{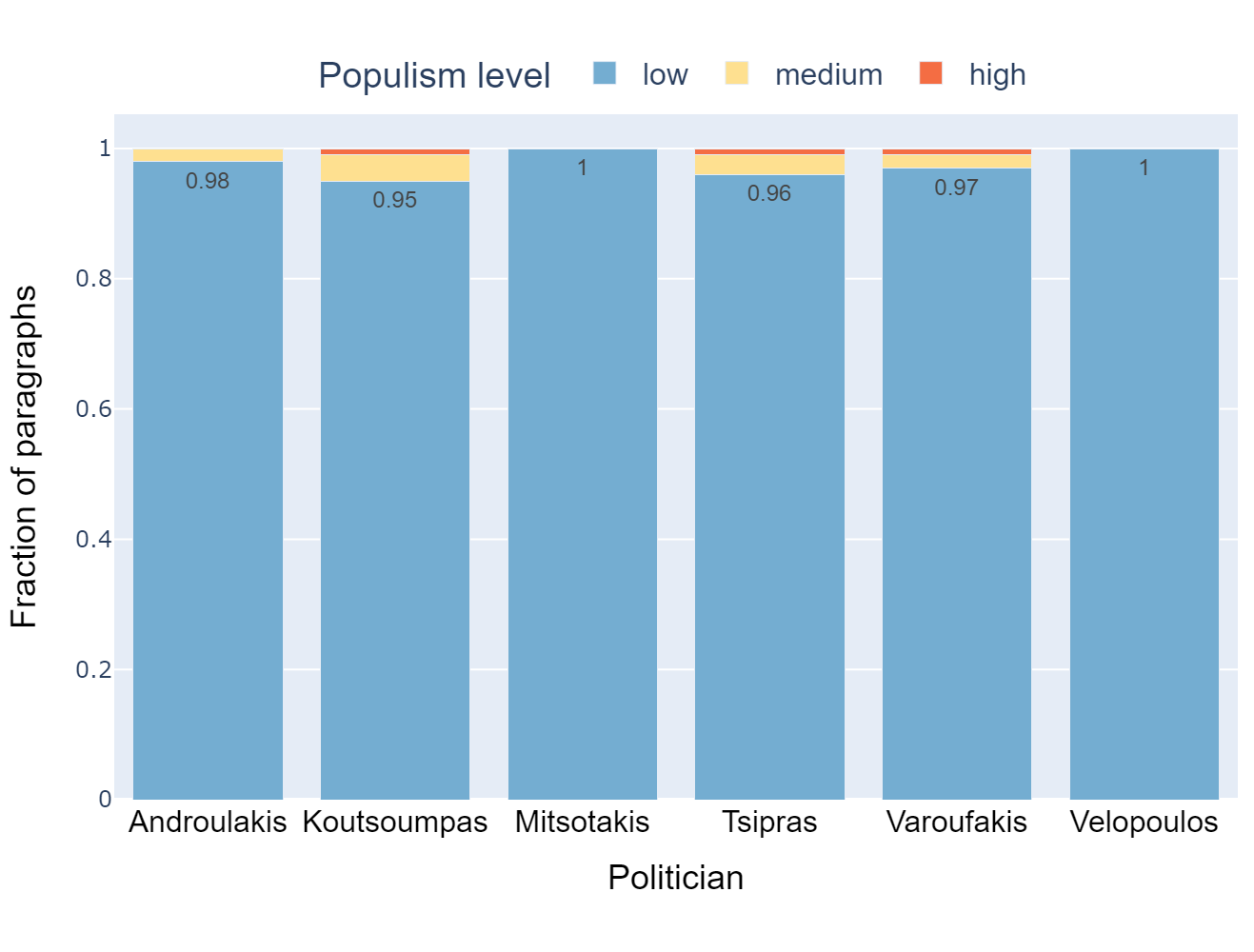}}\\
    \caption{The distribution of polarization (\ref{sfig:polarization}) and populism (\ref{sfig:populism}) in the campaign speeches within the dataset, broken down by each politician.}
    \label{fig:polarizationpopulism}
\end{figure}

For identifying polarization discursive patterns in campaign speeches, Figure \ref{sfig:polarization} presents the distribution of paragraph polarization levels for each politician, revealing clear differences. Mitsotakis stands out with the least polarized rhetoric, with 94\% of his paragraphs being zero or low in polarization and only 6\% falling into the medium and high categories, deeply connected to his strategy of promoting a positive message and avoiding criticizing his political adversaries. Koutsoumpas, in contrast, has the highest polarization, with only 78\% of his content classified as zero or low-polarized, while 16\% is medium and 6\% highly polarized, reflecting a more confrontational style. The other politicians -Tsipras, Androulakis, Varoufakis, and Velopoulos- fall between these extremes, with 86\% to 89\% of their content being zero or low-polarized, and the remainder distributed between medium and high polarization, reflecting a more balanced but varied strategy.

%The stark contrast between Mitsotakis and Koutsoumpas likely stems from Mitsotakis' use of a restrained, low-polarized style, aimed at appealing to a broader audience, while Koutsoumpas adopts a more polarized tone, reflecting his strong ideological stance. The other politicians strike a middle ground, displaying moderate levels of polarization but occasionally employing stronger rhetoric.

\begin{graycomment}
    Politicians' speeches range in polarization, with Mitsotakis being the least polarized, Koutsoumpas the most, and the others adopting a more moderate approach.
\end{graycomment}

\subsubsection*{Do politicians' speeches exhibit populism?  (populism detection)}
For identifying populism discursive patterns in campaign speeches, Figure \ref{sfig:populism} presents the distribution of paragraph populism levels for each politician, revealing a clear tendency toward zero or low populism across all six. At least 95\% of their content falls within the no/low populism category, with Mitsotakis and Velopoulos, completely avoiding the use of typical populist discourse. Androulakis, Varoufakis, Tsipras, and Koutsoumpas' pre-election speeches also fall predominantly within the no/low populism category, with 98\%, 97\%, 96\%, and 95\%, respectively. Consequently, no politician articulated a fully developed populist discourse during the two pre-election campaigns. The analysis reveals only a limited number of populist discursive patterns, primarily identified in the speeches of Koutsoumpas and Tsipras, however, this does not mean that either adopted a populist strategy. This pattern highlights the absence of a strong populist discourse during the pre-election periods, with Mitsotakis and Velopoulos particularly notable for consistently avoiding any typically populist reference.

\begin{graycomment}
    Politicians' speeches primarily exhibit zero or low levels of populism.
\end{graycomment}

\subsubsection*{Do politicians often refer to specific entities? (named entities recognition)}
\begin{figure}
\centering
\subfloat[Dataset overall]{\label{sfig:wordcloud}\includegraphics[width=.299\textwidth, height=5.5cm]{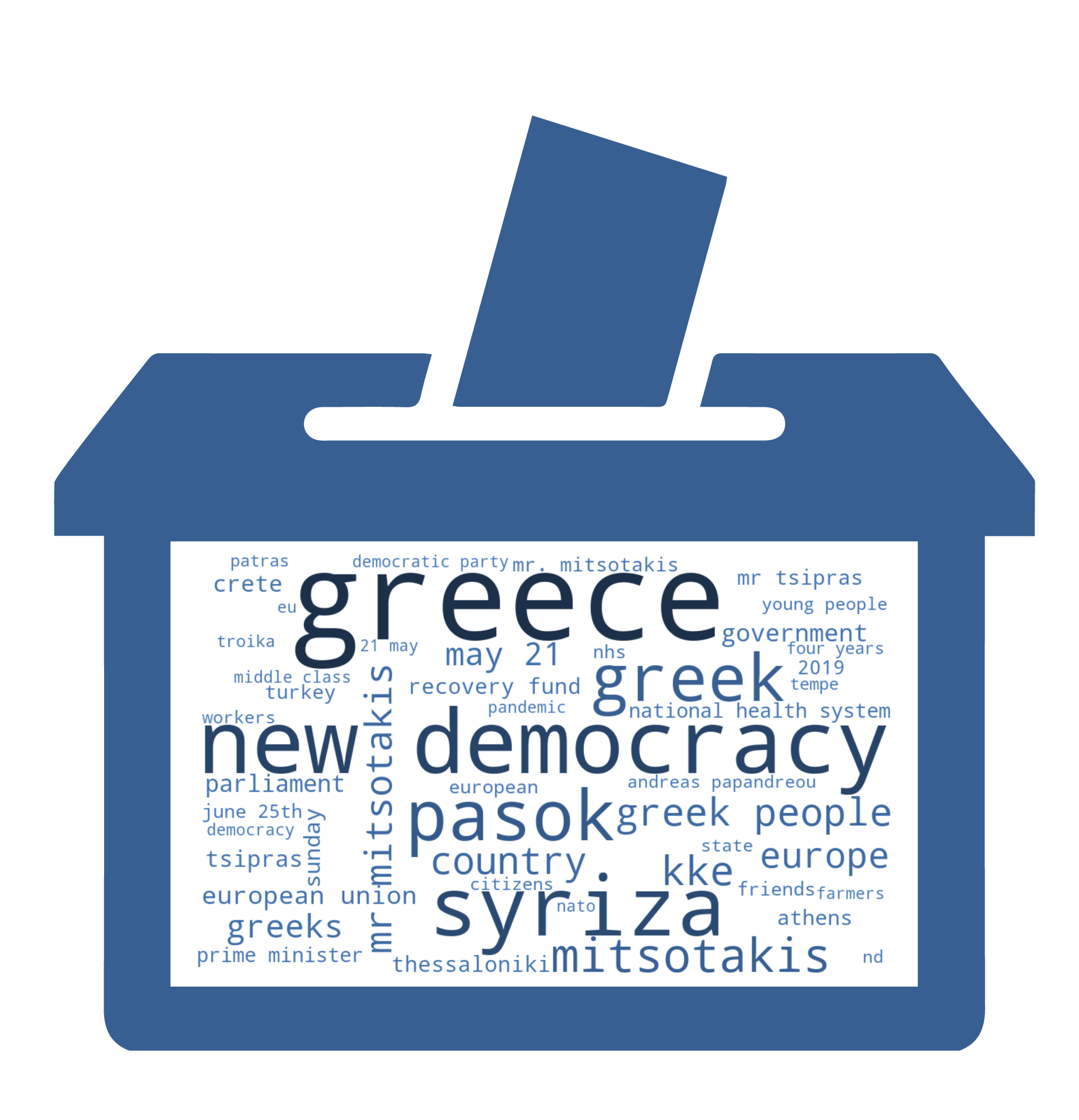}}\hfill
\subfloat[Individual politician]{\label{sfig:stackedhorizontal}\includegraphics[width=.7\textwidth]{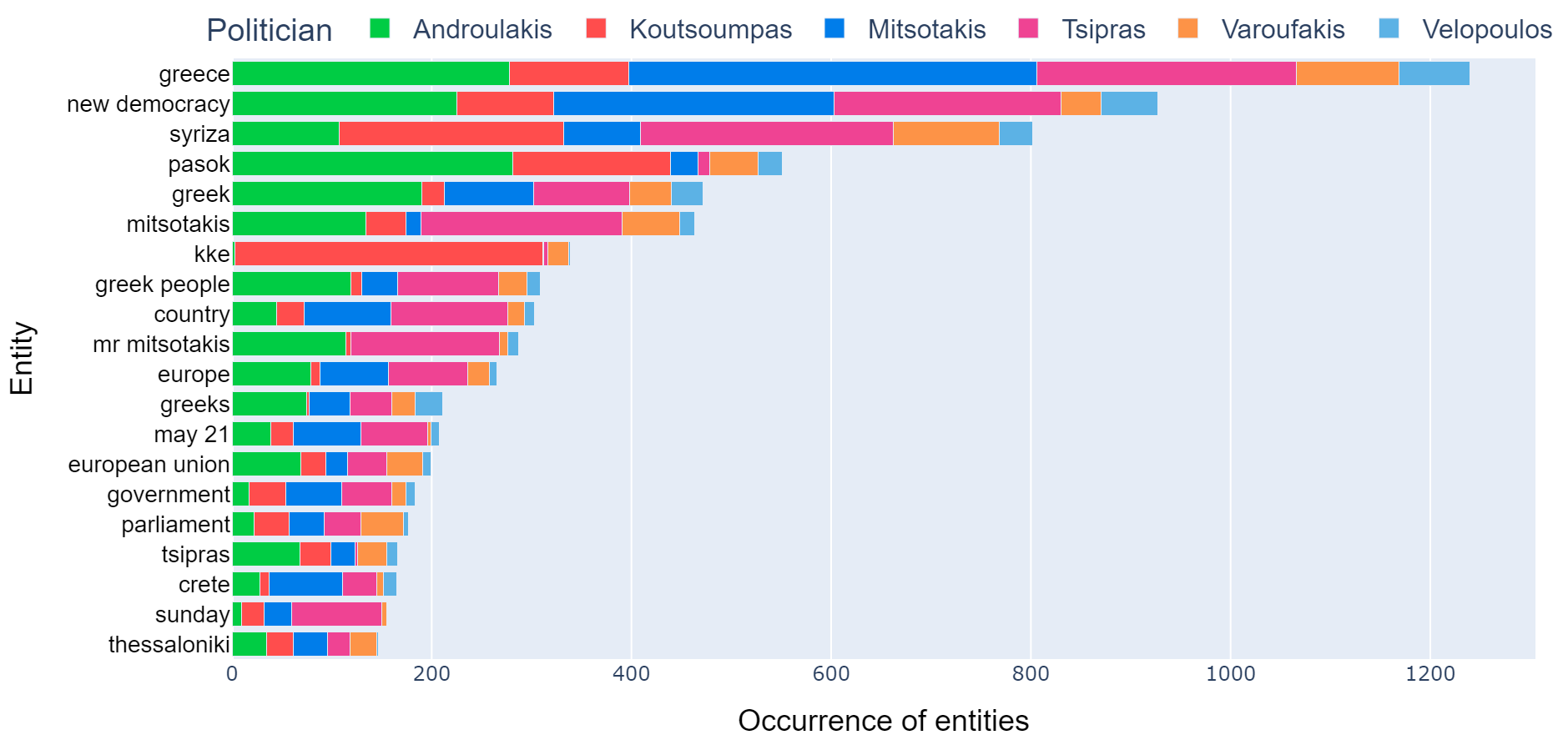}}\\
\caption{The 50 most frequently mentioned named entities across the campaign speeches of the dataset overall (\ref{sfig:wordcloud}) and the distribution of the 20 most frequent named entities break down by politician (\ref{sfig:stackedhorizontal}).}
\label{fig:promptssystem}
\end{figure}

For locating the most frequently mentioned named entities in campaign speeches, Figure \ref{sfig:wordcloud} displays a word cloud where larger words indicate a higher frequency of mentions. Prominent terms like "Greece", "democracy", and "Mitsotakis" dominate, highlighting a strong focus on the country, its political system, and current leadership. The prominence of "New Democracy," the ruling party, underscores its significant role in the discourse, while mentions of opposition parties like "SYRIZA" and "PASOK" indicate a competitive political landscape. Other frequently mentioned entities, such as "Greek people", "country", and "government", reflect a focus on national issues, while references to "Europe" and the "European Union" suggest discussions on Greece's international relations. The inclusion of "parliament", "elections", and "economy" points to the political context, driven by the upcoming elections. Finally, the campaign speeches address both long-standing concerns important to Greece, such as the "recovery fund", "national health system", "middle class", and "troika" as well as emerging issues like "Tempe" and "NATO".

Figure \ref{sfig:stackedhorizontal} compares the frequency of the most frequent named entities mentioned by each politician. All politicians frequently mention "Greece," "greek," "greek people," and "country". Additionally, "New Democracy" and "Mitsotakis," as the current prime minister and his party, along with "SYRIZA" as the main opposition, are mentioned by all politicians. The key difference lies in Mitsotakis’ emphasis on governance and Greece’s role in Europe, likely reflecting his position as head of the government. In contrast, the other politicians prioritize party-related entities, focusing on political parties and individual figures, likely indicating a more party- and criticism-centric discourse.

\begin{graycomment}
    Political campaign speeches emphasize national identity ("Greece, "greek people") and party-related entities ("New Democracy", "Mitsotakis", "SYRIZA"), while also addressing governance issues ("recovery fund", "Europe") and timely topics ("Tempe").
\end{graycomment}

\subsubsection*{How are campaign speeches characteristics related?}
\begin{figure}[htp]
    \centering
    \includegraphics[width=0.8\textwidth]{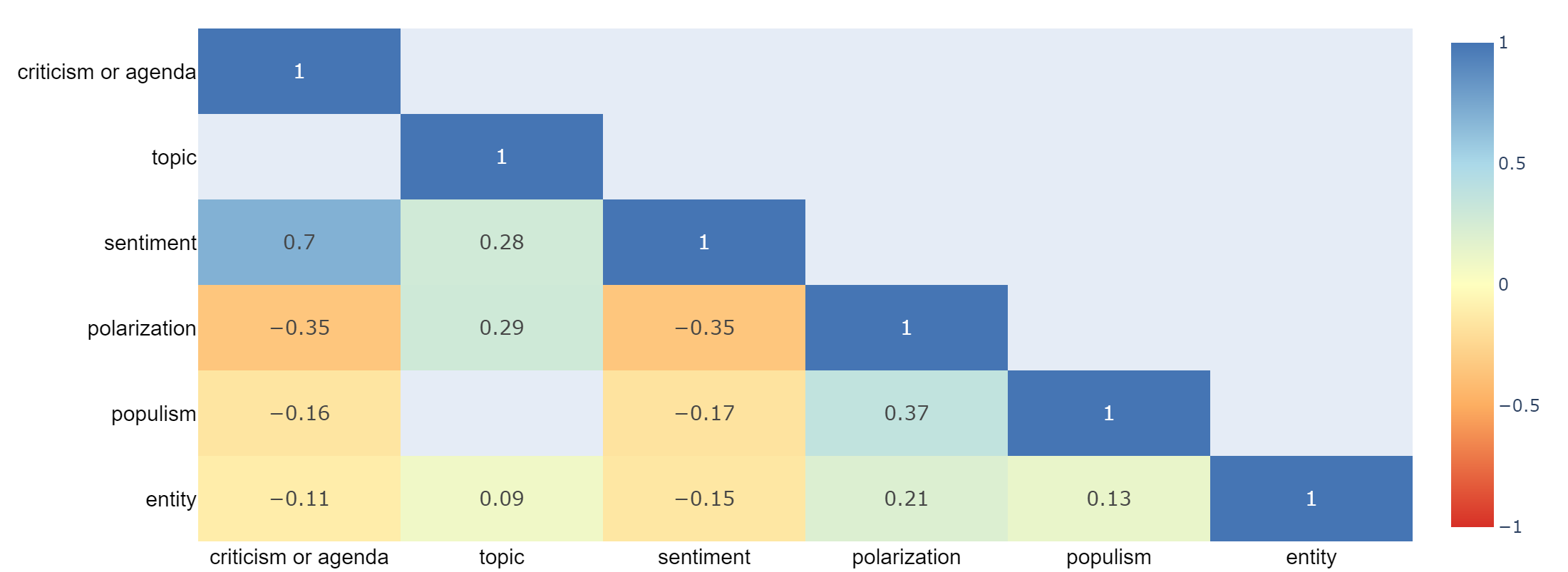}
    \caption{The correlations between the characteristics of the campaign speeches across the entire dataset.}
    \label{fig:correlations}
\end{figure}

For exploring the relationships among all the above characteristics in political campaign speeches, Figure \ref{fig:correlations} presents a matrix displaying the correlations between all variable pairs. These correlations are computed using Spearman's correlation \cite{spearman1961proof} for numerical variables and Cramér's V \cite{cramer1999mathematical} for relationships between categorical variables and between categorical and numerical variables. The correlation values range from -1 to 1, where values closer to 1 indicate a strong positive correlation, meaning both variables increase together, and values closer to -1 suggest a strong negative correlation, where one variable increases as the other decreases.

The most notable finding is the strong positive correlation (0.7) between "sentiment" and "criticism or agenda", indicating that these two characteristics tend to move together in the dataset. Specifically, as political leaders focus more on presenting their agenda, the sentiment in their speeches becomes more positive, whereas when they criticize their opponents, negative sentiment tends to dominate. Additionally, several moderate correlations are observed. Notably, there is a negative correlation between "polarization" and both "criticism or agenda" (-0.35) and "sentiment" (-0.35), suggesting that more polarized speeches are associated with negative sentiment and a reduced emphasis on agenda-driven discourse. Another moderate correlation, but positive, is seen between "populism" and "polarization" (0.37) since both discourses have a dichotomic structure and an antagonistic dimension. Lastly, two other notable correlations include the relationship between "topic" and both "polarization" (0.29) and "sentiment" (0.28), suggesting that certain topics are more likely to be associated with polarized speech and specific sentiment levels.

\begin{graycomment}
    Sentiment is strongly correlated with criticism or agenda, while there are several moderate correlations, such as between polarization and both sentiment and criticism or agenda, and between populism and polarization, among others.
\end{graycomment}

\subsection*{Limitations}
In this study for developing the \mydataset dataset, we faced several challenges that should be acknowledged:

\textit{Data sources:} %The data source is restricted to specific political speeches of the political leaders and does not include social media data, which exhibit different linguistic characteristics. As a result, the applicability of this dataset to NLP models and tasks designed for social media content analysis needs to be explored in future research. 
Part of the data sources for the speeches were in audio and video format, which were then transcribed to test. This could affect the effectiveness of tasks like sentiment analysis, which are limited to patterns in written text. Sarcasm and other emotional nuances are often conveyed through tone and intonation, which are absent in text-only data.

\textit{Translation:} At the time of annotation, the ChatGPT model available was not proficient in understanding or generating responses in Greek. This necessitated translating the content from Greek to English, but future model versions might allow for native interaction in Greek, eliminating the need for translation. While well-established automated tools were used for transcription and translation, there is always a risk of losing meaning, nuance, or local language intricacies.

\textit{Paragraph-level analysis:} The analysis was conducted at the paragraph level, which may result in missed context, particularly in complex tasks like detecting populism or polarization. These issues rely heavily on the broader context of a speech, and breaking it into paragraphs can obscure the nuances necessary for accurate analysis. This limitation is especially noticeable in ChatGPT annotations, where the lack of surrounding context can lead to misinterpretation.

\textit{ChatGPT prompting:} At the time of the project, careful selection of prompts was crucial for the quality of responses. Newer models of ChatGPT are more effective and robust, eliminating the need for elaborate prompt fine-tuning.

\section*{Usage Notes}
We provide the \mydataset dataset with full open access to "enable third parties to access, mine, exploit, reproduce and disseminate (free of charge for any user) this research data" (\url{https://ec.europa.eu/research/participants/docs/h2020-funding-guide/cross-cutting-issues/open-access-data-management/open-access_en.htm}). We encourage everyone who uses the data to abide by the following code of conduct:
\begin{itemize}[itemsep=0.05em, labelsep=0.05em]
    \item abide by the guidelines for ethical research as described in the ACM Code of Ethics and Professional Conduct~\cite{gotterbarn2018acm};
    \item report any misuse, intentional or not, to the corresponding authors by mail within a reasonable time;
    \item analyze the \mydataset dataset ethically, ensuring that it does not mislead, misrepresent, or contribute to biased or false narratives;
    \item include a proper reference on all publications or other communications resulting from using the LifeSnaps data.
\end{itemize}

\section*{Code Availability}
All code for reading, processing, and exploring the data is made openly available at (\url{https://github.com/Datalab-AUTH/AgoraSpeech-EDA}).

\bibliography{main}

\section*{Acknowledgements} 
We would like to thank Nota Vafea, Katerina Voutsina, Stefania Ibrishimova, Athina Thanasi, Chrysoula Marinou, and Georgios Schinas (iMEdD) for their contributions to the human data annotation process. We also thank Christos Nomikos, Nikos Sarantos (iMEdD), and Dimitrios-Panteleimon Giakatos (Datalab) for their IT support and software development of the online tool, as well as Anatoli Stavroulopoulou for cross-checking the political speeches' translations.

% The 'Acknowledgements' statement should contain text acknowledging non-author contributors. Acknowledgements should be brief, and should not include thanks editors or effusive comments. Grant or contribution numbers may be acknowledged.

\section*{Author Contributions Statement}
P.S., S.K., I.D., T.T., K.K., A.G. and A.V. conceived and designed the study, P.S., S.K., I.D., T.T., K.K., and A.G. conducted the study, and P.S., S.K., I.D., E.P., S.Y., and F.K. processed, structured and analyzed the data and results. All authors reviewed the manuscript.

\section*{Competing Interests} 
The authors declare no competing interests.

\end{document}